%% file: paper.tex
\documentclass[twocolumn]{article}
\usepackage{geometry}
\usepackage{microtype}
\usepackage{graphicx}
\usepackage{booktabs} % for professional tables

\usepackage{hyperref}

\usepackage{amsmath}
\usepackage{amssymb}
\usepackage{mathtools}
\usepackage{amsthm}

\usepackage[capitalize,noabbrev]{cleveref}

\theoremstyle{plain}

\theoremstyle{definition}

\theoremstyle{remark}

\usepackage[disable,textsize=tiny]{todonotes}
\input{math_commands.tex}

\usepackage{url}
\usepackage{graphicx}
\usepackage{xcolor}
\usepackage{colortbl}
\usepackage{import}
\usepackage{multirow}
\usepackage{wrapfig}
\usepackage{pdflscape}
\usepackage{float}
\usepackage{subcaption}
\usepackage{authblk}
\usepackage{natbib}
\usepackage{libertine}
\usepackage{fancyhdr}
\usepackage{pifont}
\newcommand{\cmark}{\ding{51}}%
\newcommand{\xmark}{\ding{55}}%
\title{HYVE: Hybrid Vertex Encoder for Neural Distance Fields}

\author[1]{Stefan R. Jeske}
\author[2]{Jonathan Klein}
\author[2]{Dominik Michels}
\author[1]{Jan Bender}

\affil[1]{Computer Animation, RWTH Aachen University, Aachen, Germany}
\affil[2]{Computational Sciences Group, KAUST, Thuwal, Saudi Arabia}

\date{}

\begin{document}

\twocolumn[
    \begin{@twocolumnfalse}
        \maketitle

        \begin{abstract}
Neural shape representation generally refers to representing 3D geometry using neural networks, e.g., computing a signed distance or occupancy value at a specific spatial position. 
In this paper we present a neural-network architecture suitable for accurate encoding of 3D shapes in a single forward pass.
Our architecture is based on a multi-scale hybrid system incorporating graph-based and voxel-based components, as well as a continuously differentiable decoder. 
The hybrid system includes a novel way of voxelizing point-based features in neural networks, which we show can be used in combination with oriented point-clouds to obtain smoother and more detailed reconstructions.
Furthermore, our network is trained to solve the eikonal equation and only requires knowledge of the zero-level set for training and inference. 
This means that in contrast to most previous shape encoder architectures, our network is able to output valid signed distance fields without explicit prior knowledge of non-zero distance values or shape occupancy. 
It also requires only a single forward-pass, instead of the latent-code optimization used in auto-decoder methods.
We further propose a modification to the loss function in case that surface normals are not well defined, e.g., in the context of non-watertight surfaces and non-manifold geometry, resulting in an unsigned distance field.
Overall, our system can help to reduce the computational overhead of training and evaluating neural distance fields, as well as enabling the application to difficult geometry. 
We finally demonstrate the efficacy, generalizability and scalability of our method on variety of datasets, containing temporally-deforming shapes and single/multi-class objects, obtained from physical simulations, 3D scans and hand-made models.
Our method consistently outperforms both well-established baselines as well as more recent methods on the surface reconstruction task, while being more computationally efficient and requiring fewer parameters.
        \end{abstract}
        \vskip 0.3in
    \end{@twocolumnfalse}
]
\begin{figure*}[!h]
   \def\svgwidth{\linewidth}
   \fontsize{8}{9}\selectfont
   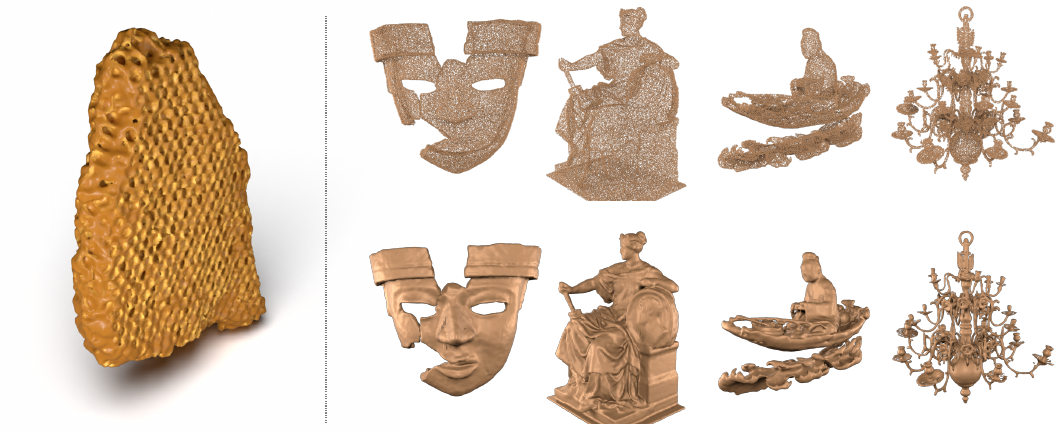
   \caption{The reconstruction of a 3D scan of a beehive (left) and comparisons of input points and our respective reconstructions on the Objaverse dataset (right) \cite{deitkeObjaverseUniverseAnnotated2023}. 
   These examples show the capability of our model to encode a large amount of detail in a single forward pass, using only oriented point clouds as input.}
   \label{fig:teaser}
\end{figure*}
% 
% Keywords section (optional)

% > 510.295pt.
% \showthe\textwidth
%                         
% > 243.14749pt.
% \showthe\linewidth

\input{sections/1-introduction.tex}

\input{sections/2-related-work.tex}
\input{sections/3-method.tex}
\input{sections/4-results.tex}
\input{sections/5-conclusion.tex}

\bibliography{bibliography}

%%%%%%%%%%%%%%%%%%%%%%%%%%%%%%%%%%%%%%%%%%%%%%%%%%%%%%%%%%%%%%%%%%%%%%%%%%%%%%%
%%%%%%%%%%%%%%%%%%%%%%%%%%%%%%%%%%%%%%%%%%%%%%%%%%%%%%%%%%%%%%%%%%%%%%%%%%%%%%%
% APPENDIX
%%%%%%%%%%%%%%%%%%%%%%%%%%%%%%%%%%%%%%%%%%%%%%%%%%%%%%%%%%%%%%%%%%%%%%%%%%%%%%%
%%%%%%%%%%%%%%%%%%%%%%%%%%%%%%%%%%%%%%%%%%%%%%%%%%%%%%%%%%%%%%%%%%%%%%%%%%%%%%%
% \newpage
% \appendix
% \onecolumn

% \input{sections/appendix.tex}

%%%%%%%%%%%%%%%%%%%%%%%%%%%%%%%%%%%%%%%%%%%%%%%%%%%%%%%%%%%%%%%%%%%%%%%%%%%%%%%
%%%%%%%%%%%%%%%%%%%%%%%%%%%%%%%%%%%%%%%%%%%%%%%%%%%%%%%%%%%%%%%%%%%%%%%%%%%%%%%

\end{document}

%% file: math_commands.tex
\usepackage{amsmath,amsfonts,bm}

\def\eqref#1{equation~\ref{#1}}
\def\Eqref#1{Equation~\ref{#1}}
\def\1{\bm{1}}

\def\vf{{\bm{f}}}

\def\vn{{\bm{n}}}

\def\vx{{\bm{x}}}

\DeclareMathAlphabet{\mathsfit}{\encodingdefault}{\sfdefault}{m}{sl}
\SetMathAlphabet{\mathsfit}{bold}{\encodingdefault}{\sfdefault}{bx}{n}

\def\gL{{\mathcal{L}}}

\newcommand{\R}{\mathbb{R}}

%% file: figures/teaser.pdf_tex
%% Creator: Inkscape 1.3 (0e150ed, 2023-07-21), www.inkscape.org
%% Accompanies image file 'teaser.pdf' (pdf, eps, ps)
\begingroup%
  \makeatletter%
  \providecommand\color[2][]{%
    \errmessage{(Inkscape) Color is used for the text in Inkscape, but the package 'color.sty' is not loaded}%
    \renewcommand\color[2][]{}%
  }%
  \providecommand\transparent[1]{%
    \errmessage{(Inkscape) Transparency is used (non-zero) for the text in Inkscape, but the package 'transparent.sty' is not loaded}%
    \renewcommand\transparent[1]{}%
  }%
  \providecommand\rotatebox[2]{#2}%
  \newcommand*\fsize{\dimexpr\f@size pt\relax}%
  \newcommand*\lineheight[1]{\fontsize{\fsize}{#1\fsize}\selectfont}%
  \ifx\svgwidth\undefined%
    \setlength{\unitlength}{510.29501343bp}%
    \ifx\svgscale\undefined%
      \relax%
    \else%
      \setlength{\unitlength}{\unitlength * \real{\svgscale}}%
    \fi%
  \else%
    \setlength{\unitlength}{\svgwidth}%
  \fi%
  \global\let\svgwidth\undefined%
  \global\let\svgscale\undefined%
  \makeatother%
  \begin{picture}(1,0.4056477)%
    \lineheight{1}%
    \setlength\tabcolsep{0pt}%
    \put(0,0){\includegraphics[width=\unitlength,page=1]{teaser.pdf}}%
    \put(0.33414701,0.10059226){\color[rgb]{0,0,0}\rotatebox{90}{\makebox(0,0)[t]{\lineheight{1.25}\smash{\begin{tabular}[t]{c}Our Reconstruction\end{tabular}}}}}%
    \put(0.33431925,0.31151258){\color[rgb]{0,0,0}\rotatebox{90}{\makebox(0,0)[t]{\lineheight{1.25}\smash{\begin{tabular}[t]{c}Input Points\end{tabular}}}}}%
    \put(0,0){\includegraphics[width=\unitlength,page=2]{teaser.pdf}}%
  \end{picture}%
\endgroup%

%% file: sections/1-introduction.tex
\section{Introduction}
\label{sec:introduction}

Algorithms processing 3D geometric data have become omnipresent and an integral part of many systems. 
These include for example the systems evaluating Lidar sensor data, game engine processing, visualizing 3D assets, and physical simulation used in engineering prototypes.
In recent years, deep learning methods have been increasingly investigated to assist in solving problems pertaining to 3D geometry.

In particular, neural shape representation deals with using neural networks to predict shape occupancies or surface distances at arbitrary spatial coordinates.
Recent works have shown the ability to capture intricate details of 3D geometry with ever increasing fidelity \citep{wangGeometryconsistentNeuralShape2022,longNeuralUDFLearningUnsigned2023}.
However, a significant number of such works employ an auto-decoder based architecture, which requires solving an optimization problem when representing new geometry. 
Additionally, the auto-decoder still has to be evaluated for all query points individually, which can become very costly when evaluating these systems for high-resolution reconstruction \citep{xieNeuralFieldsVisual2022}.
Finally, many of these methods also require annotated and densely sampled ground truth data.
Meta learning approaches, e.g., by \citet{sitzmannMetaSDFMetalearningSigned2020,ouasfiFewZeroLevel2022}, mitigate this problem, but also have to run several iterations of gradient descent to specialize the network for each new model before inference.
Encoder-decoders can instead encode the shape in a single forward pass \citep{chibaneImplicitFunctionsFeature2020,boulchPOCOPointConvolution2022}, and typically employ computationally cheaper decoder networks for evaluating query points. 
Nevertheless, these approaches also often require elaborate data preprocessing pipelines, and rely on labeled training data.

One of our core motivations for this work was to develop a neural distance field based method that could be used for signed distance computation of rigid and deforming objects.
Both are represented as points and the latter can change often, requiring frequent recomputation of the distance field.
As a result, we wanted to keep the decoder network computationally efficient and instead of latent optimization we wanted to develop an encoder capable of generating fast and accurate latent codes for this decoder network.

Therefore, we propose an end-to-end learnable encoder-decoder system, that is not bound by previous data preprocessing constraints and can be trained using only the zero-level set, i.e., surface samples as labeled data. 
This kind of training was previously introduced for auto-decoders \citep{sitzmannImplicitNeuralRepresentations2020}, and is enabled by using the eikonal equation as a training target.
Similar approaches have been attempted for encoders by \citep{atzmonSALSignAgnostic2020,atzmonSALDSignAgnostic2020}, using global encoders augmented by latent-optimization at test-time.
Yet to the best of our knowledge, this kind of training is not well-established using encoders without latent-optimization.

We summarize our contributions as follows:
\begin{itemize}
	\item We derive an encoder architecture for representing 3D shapes. 
The key component of this encoder is the hybrid and interleaved execution of graph-level convolutions and 3D grid convolutions, as well as back-and-forth feature projections.
Our motivation for this hybrid approach is the ability to accurately \emph{extract} information from surface-points and to efficiently \emph{process} information in grid (Euclidean) space.  
	\item We introduce a novel way of voxelizing point-features within network architectures, in which the pooling operator is replaced by projection of the entire ``feature-field'' onto the grid.
We show that this results in less-noisy and more detailed surfaces, especially when oriented point-clouds are used as input.
	\item We show that the accuracy of our architecture is intuitively controllable. 
Using a model with as little as 38K parameters (including the decoder) can already achieve excellent visual quality while being very fast to evaluate.
This makes it useful for practical applications within resource constrained computing platforms.
	\item We show the feasibility of training encoder-decoder networks on the eikonal equation for high-fidelity 3D shape encoding. 
We also propose a simple yet effective modification to the loss function that can gracefully handle poorly oriented surface normals in the training data, e.g., caused by non-manifold or non-watertight geometry. 
\end{itemize}

In the evaluation we show, that we are able to reconstruct better quality surfaces than other state-of-the-art methods.
Sample reconstructions of our method are shown in Figure~\ref{fig:teaser}.

%% file: sections/2-related-work.tex
\section{Related Work}
\label{sec:related-work}

\begin{figure*}[t]
\fontsize{8}{9}\selectfont
\centering
\def\svgwidth{\linewidth}
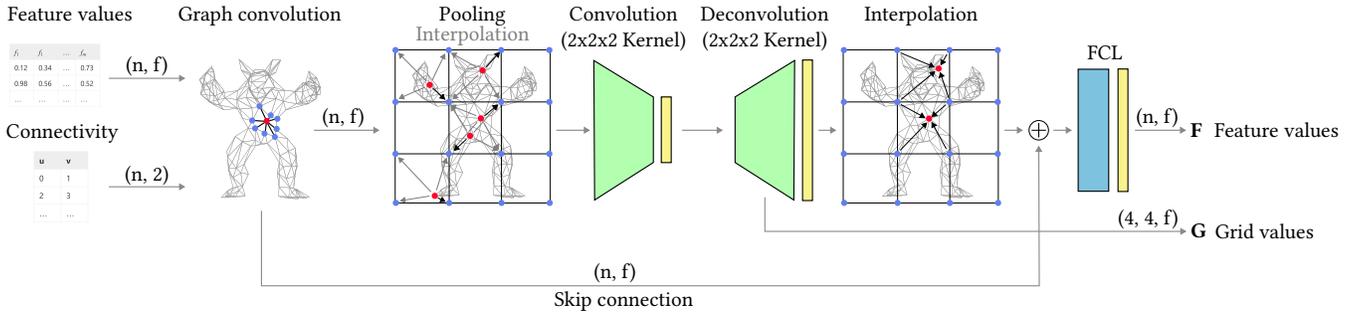
\caption{\label{fig:grid-block}Convolution block that extracts features for a specific grid resolution.
For clarity of illustration, a 2D rather than 3D grid is shown here.
The input is a set of vertices (with position / feature data) and edges (encoded as vertex indices).
The + denotes element-wise vector addition.
The block has two outputs, feature values on the vertices and grid values for each grid cell.
For all resolutions, a $2\times2$ convolution kernel is used.
$n$: number of vertices.
$f$: number of features (on the first level, the features are the spatial coordinate of each vertex).
}
\end{figure*}
Neural Fields have become an integral part of research in geometric deep learning, with hundreds of papers published in recent years.
A comprehensive overview is given by \citet{xieNeuralFieldsVisual2022}.
One of the seminal works on deep learning for unstructured data was the introduction of PointNet \citep{qiPointNetDeepLearning2017}. 
From today's perspective, one of the major limitations of this work is the difficulty of learning high-frequency functions from low-dimensional data \citep{xuTrainingBehaviorDeep2019,rahamanSpectralBiasNeural2019}.
The solution to the problem is addressed by more recent approaches such as NeRFs \citep{mildenhallNeRFRepresentingScenes2020} and Fourier Feature Networks \citep{tancikFourierFeaturesLet2020}.
In essence, the idea is to use positional embeddings, inspired by the embeddings proposed by \citet{vaswaniAttentionAllYou2017} for transformer networks.
These embeddings compute a mapping from low dimensional positional information (typically 2D or 3D) into higher dimensional spaces using a specific number of Fourier basis functions \citep{tancikFourierFeaturesLet2020}.
A concurrent work shows that using periodic activation functions inside an MLP also significantly improves reconstruction quality and surface detail \citep{sitzmannImplicitNeuralRepresentations2020},
a single layer of which can again be seen as a kind of positional encoding \citep{xieNeuralFieldsVisual2022}.
Subsequent works improve the usage of positional encodings, e.g., by controlling the frequency through a feedback loop \citep{hertzSAPESpatiallyAdaptiveProgressive2021} or modulating the periodic activations using a separate ReLU activated MLP \citep{mehtaModulatedPeriodicActivations2021}.
Other benefits of using periodic activations are the ability to better learn high-frequency mappings and the continuous differentiability of these activations which is useful for evaluating network derivatives as training targets \citep{raissiPhysicsinformedNeuralNetworks2019,wangGeometryconsistentNeuralShape2022,chenCROMContinuousReducedorder2023}.

There are many approaches for representing 3D shapes using neural networks.
For clarity of exposition we will classify them into global and local methods.

\paragraph{Global methods} These do not make use of geometric structures in the network itself, and can generally be used irrespective of the discretized representation of the geometry.
Typically auto-decoder methods, in which the latent representation is optimized during training and testing, are in this category \citep{parkDeepSDFLearningContinuous2019,meschederOccupancyNetworksLearning2019,sitzmannImplicitNeuralRepresentations2020,groppImplicitGeometricRegularization2020,atzmonSALSignAgnostic2020,atzmonSALDSignAgnostic2020,mehtaModulatedPeriodicActivations2021,wangGeometryconsistentNeuralShape2022}.
The network can then be queried using both the latent feature and a 3D coordinate, to evaluate either a distance metric, or an occupancy value.
Meta learning approaches also fall into this category.
A few iterations of gradient descent are used to specialize the weights of a generalized network to a new shape \citep{sitzmannMetaSDFMetalearningSigned2020,ouasfiFewZeroLevel2022}.
An approach that has both discretization-dependent and independent components was presented by \citet{chenCROMContinuousReducedorder2023}, where the discretization-dependent encoder is typically discarded during inference.
The amount of encoded details by these methods is naturally bounded by the number of network weights.
It has also been shown that using pooling-based set encoders for global conditioning frequently underfits the data \citep{buterezGraphNeuralNetworks2022}.

\paragraph{Local methods} This kind of method typically relies on using spatial structures within the network itself for the extraction of meaningful information \citep{lombardiNeuralVolumesLearning2019,pengConvolutionalOccupancyNetworks2020,chabraDeepLocalShapes2020,jiangLocalImplicitGrid2020,chibaneImplicitFunctionsFeature2020,tangSAConvONetSignAgnosticOptimization2021,boulchPOCOPointConvolution2022,zhang3DILGIrregularLatent2022,zhang3DShape2VecSet3DShape2023}.
This has proven to be a valuable approach, since it is quite difficult for neural networks to encode the high-frequency functions needed to represent detailed fields in 3D.
Previous works make effective use of discretized structures, e.g., point-clouds, meshes or voxel-grids as either inputs or outputs \citep{qiPointNetDeepLearning2017,hertzSAPESpatiallyAdaptiveProgressive2021,takikawaNeuralGeometricLevel2021,buterezGraphNeuralNetworks2022}. For encoder-type methods, extracting local features has been shown to improve network performance over global ones.
Nevertheless, many works focus on either purely point-based or grid-based approaches with prior voxelization of input points.
An early work by \citet{liuPointVoxelCNNEfficient2019} explores the use of hybrid point-voxel architectures for object segmentation and classification and shows promising results in terms of accuracy and efficiency.
Therefore, we found hybrid methods to be a promising direction for devising a 3D reconstruction neural network, where these properties are of high importance.
In contrast to previous work, we expand the capabilities of hybrid approaches to the unsupervised 3D reconstruction task and explore important design decisions.
This includes details on network construction and the often overlooked point-to-grid feature transfer methods to maintain consistency between grid and points.
We also show the scalability to high-resolution point clouds with up to 100k points during inference.

\paragraph{Unsupervised shape encoding} A number of recent works have investigated the ability to train networks without direct supervision in order to learn shape representation \citep{atzmonSALDSignAgnostic2020,atzmonSALSignAgnostic2020,sitzmannImplicitNeuralRepresentations2020,groppImplicitGeometricRegularization2020,maNeuralPullLearningSigned2021,pengShapePointsDifferentiable2021,chibaneNeuralUnsignedDistance2020,longNeuralUDFLearningUnsigned2023}.
Most of these works focus on auto-decoders and generally use optimization of latent codes during inference, while others ``overfit'' small networks to specific shapes \citep{maNeuralPullLearningSigned2021}.
Some also make use of learning unsigned distance fields, which enables training on non-manifold geometry, but makes reconstruction more difficult. 
While \citet{tangSAConvONetSignAgnosticOptimization2021} also use sign-agnostic optimization of occupancy fields, they still require ground-truth occupancy values for pre-training.
Notably, \citet{groppImplicitGeometricRegularization2020} introduced the formulation of the unsupervised eikonal loss, which was further refined in the work of \citet{sitzmannImplicitNeuralRepresentations2020}.
In our work we extend this loss to improve training on data with inconsistent normal orientations.

%% file: figures/grid_level_block.pdf_tex
%% Creator: Inkscape 1.3 (0e150ed, 2023-07-21), www.inkscape.org
%% Accompanies image file 'grid_level_block.pdf' (pdf, eps, ps)
\begingroup%
  \makeatletter%
  \providecommand\color[2][]{%
    \errmessage{(Inkscape) Color is used for the text in Inkscape, but the package 'color.sty' is not loaded}%
    \renewcommand\color[2][]{}%
  }%
  \providecommand\transparent[1]{%
    \errmessage{(Inkscape) Transparency is used (non-zero) for the text in Inkscape, but the package 'transparent.sty' is not loaded}%
    \renewcommand\transparent[1]{}%
  }%
  \providecommand\rotatebox[2]{#2}%
  \newcommand*\fsize{\dimexpr\f@size pt\relax}%
  \newcommand*\lineheight[1]{\fontsize{\fsize}{#1\fsize}\selectfont}%
  \ifx\svgwidth\undefined%
    \setlength{\unitlength}{510.29501343bp}%
    \ifx\svgscale\undefined%
      \relax%
    \else%
      \setlength{\unitlength}{\unitlength * \real{\svgscale}}%
    \fi%
  \else%
    \setlength{\unitlength}{\svgwidth}%
  \fi%
  \global\let\svgwidth\undefined%
  \global\let\svgscale\undefined%
  \makeatother%
  \begin{picture}(1,0.22787406)%
    \lineheight{1}%
    \setlength\tabcolsep{0pt}%
    \put(0,0){\includegraphics[width=\unitlength,page=1]{grid_level_block.pdf}}%
    \put(0.44962528,0.02695024){\color[rgb]{0,0,0}\makebox(0,0)[t]{\lineheight{1.25}\smash{\begin{tabular}[t]{c}(n, f)\end{tabular}}}}%
    \put(0.45596098,0.21798681){\color[rgb]{0,0,0}\makebox(0,0)[t]{\lineheight{1.25}\smash{\begin{tabular}[t]{c}Convolution\\(2x2x2 Kernel)\end{tabular}}}}%
    \put(0.56066612,0.21798681){\color[rgb]{0,0,0}\makebox(0,0)[t]{\lineheight{1.25}\smash{\begin{tabular}[t]{c}Deconvolution\\(2x2x2 Kernel)\end{tabular}}}}%
    \put(0.34399086,0.21798681){\color[rgb]{0,0,0}\makebox(0,0)[t]{\lineheight{1.25}\smash{\begin{tabular}[t]{c}Pooling\end{tabular}}}}%
    \put(0.34416562,0.20292513){\color[rgb]{0.49411765,0.49411765,0.49411765}\makebox(0,0)[t]{\lineheight{1.25}\smash{\begin{tabular}[t]{c}Interpolation\end{tabular}}}}%
    \put(0.67651974,0.21798681){\color[rgb]{0,0,0}\makebox(0,0)[t]{\lineheight{1.25}\smash{\begin{tabular}[t]{c}Interpolation\end{tabular}}}}%
    \put(0.24832451,0.141383){\color[rgb]{0,0,0}\makebox(0,0)[t]{\lineheight{1.25}\smash{\begin{tabular}[t]{c}(n, f)\end{tabular}}}}%
    \put(0,0){\includegraphics[width=\unitlength,page=2]{grid_level_block.pdf}}%
    \put(0.18636219,0.21798681){\makebox(0,0)[t]{\lineheight{1.25}\smash{\begin{tabular}[t]{c}Graph convolution\end{tabular}}}}%
    \put(0,0){\includegraphics[width=\unitlength,page=3]{grid_level_block.pdf}}%
    \put(0.86722951,0.06635458){\color[rgb]{0,0,0}\makebox(0,0)[rt]{\lineheight{1.25}\smash{\begin{tabular}[t]{r}(4, 4, f)\end{tabular}}}}%
    \put(0,0){\includegraphics[width=\unitlength,page=4]{grid_level_block.pdf}}%
    \put(0.10265538,0.1799703){\color[rgb]{0,0,0}\makebox(0,0)[t]{\lineheight{1.25}\smash{\begin{tabular}[t]{c}(n, f)\end{tabular}}}}%
    \put(-0.00036853,0.21798681){\color[rgb]{0,0,0}\makebox(0,0)[lt]{\lineheight{1.25}\smash{\begin{tabular}[t]{l}Feature values\end{tabular}}}}%
    \put(-0.00157947,0.13046886){\color[rgb]{0,0,0}\makebox(0,0)[lt]{\lineheight{1.25}\smash{\begin{tabular}[t]{l}Connectivity\end{tabular}}}}%
    \put(0.8939393,0.13140869){\color[rgb]{0,0,0}\makebox(0,0)[lt]{\lineheight{1.25}\smash{\begin{tabular}[t]{l}Feature values\end{tabular}}}}%
    \put(0.89447942,0.05625264){\color[rgb]{0,0,0}\makebox(0,0)[lt]{\lineheight{1.25}\smash{\begin{tabular}[t]{l}Grid values\end{tabular}}}}%
    \put(0.85150794,0.141383){\color[rgb]{0,0,0}\makebox(0,0)[t]{\lineheight{1.25}\smash{\begin{tabular}[t]{c}(n, f)\end{tabular}}}}%
    \put(0,0){\includegraphics[width=\unitlength,page=5]{grid_level_block.pdf}}%
    \put(0.10265538,0.09879797){\color[rgb]{0,0,0}\makebox(0,0)[t]{\lineheight{1.25}\smash{\begin{tabular}[t]{c}(n, 2)\end{tabular}}}}%
    \put(0,0){\includegraphics[width=\unitlength,page=6]{grid_level_block.pdf}}%
    \put(0.88229717,0.05649408){\color[rgb]{0,0,0}\makebox(0,0)[t]{\lineheight{1.25}\smash{\begin{tabular}[t]{c}\textbf{G}\end{tabular}}}}%
    \put(0.88067683,0.13152308){\color[rgb]{0,0,0}\makebox(0,0)[t]{\lineheight{1.25}\smash{\begin{tabular}[t]{c}\textbf{F}\end{tabular}}}}%
    \put(0.81284803,0.18902144){\color[rgb]{0,0,0}\makebox(0,0)[t]{\lineheight{1.25}\smash{\begin{tabular}[t]{c}FCL\end{tabular}}}}%
    \put(0.4560976,0.00569054){\makebox(0,0)[t]{\lineheight{1.25}\smash{\begin{tabular}[t]{c}Skip connection\end{tabular}}}}%
    \put(0,0){\includegraphics[width=\unitlength,page=7]{grid_level_block.pdf}}%
  \end{picture}%
\endgroup%

%% file: sections/3-method.tex
\section{Zero-Level-Set Encoder}
\label{sec:method}

\begin{figure*}[t]
\fontsize{8}{9}\selectfont
\def\svgwidth{\linewidth}
\centering
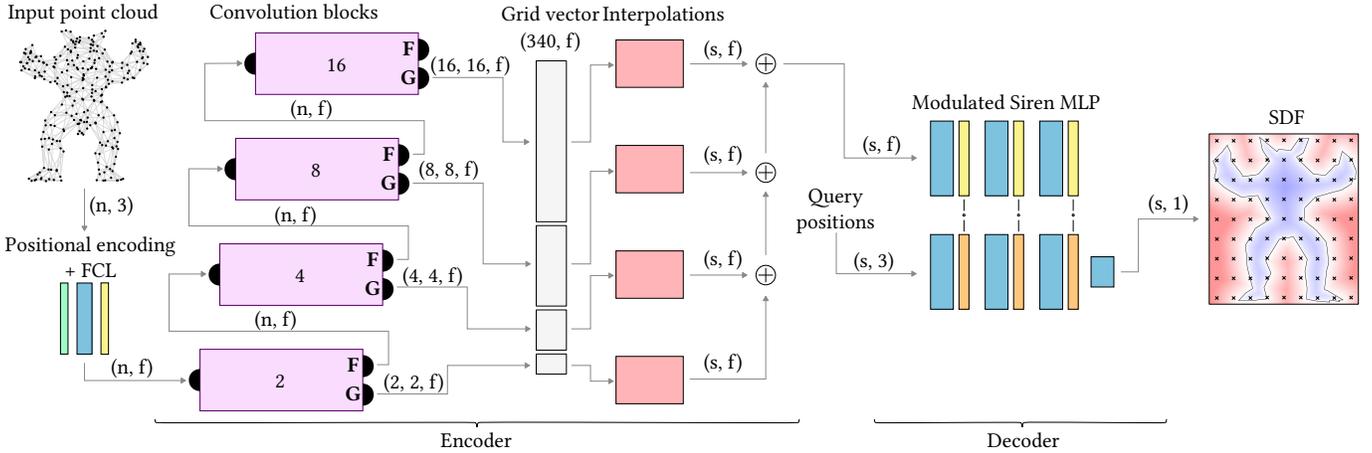
\caption{The encoder-decoder architecture of our network.
The encoder computes vertex and volumetric features at multiple resolutions.
By passing the feature vector through the convolution blocks, neighbor information is collected.
The implementation of the convolution blocks is shown in Figure~\ref{fig:grid-block}.
After the last block, the vertex feature vector is discarded.
The + denotes element-wise vector addition.
$n$: number of vertices.
$f$: number of features.
$s$: number of SDF sample points.
}
\label{fig:architecture}
\end{figure*}

In the following we will present our encoder-decoder architecture, including our \emph{convolution block} for the encoder, the decoder structure, the loss function along with our modification to support non-manifold geometries, and the overall training process.
We will also propose a new way to transfer information from points to voxels within neural networks, which is inspired by hybrid Eulerian-Lagrangian physical simulation methods.

Our inputs are surface point-clouds of 3D objects, given by a set of vertices $\mathcal{V}=\left\{ V\in\mathbb{R}^{3}\right\}$.
In order to use graph convolutions, we create edges $\mathcal{E}=\left\{ E\in\left(\mathbb{N}\times\mathbb{N}\right)\right\}$ between vertices, using e.g. k-nearest-neighbor (k-NN) or radius graph connectivity.
Within the network, the surface points store abstract $f$-dimensional feature vectors ($V^{f}\in\mathbb{R}^{f}$), rather than 3D coordinates.
This input representation also allows for utilization of known point connectivity, e.g., for partially meshed point cloud inputs or triangle soups. 

\subsection{Encoder-Decoder Architecture}
\label{sec:enc-dec}

\paragraph{Convolution block} We introduce our hybrid point-grid convolution block in Figure~\ref{fig:grid-block} as the main building block for our encoder.
This is in contrast to many previous encoder approaches that use either only point data \citep{qiPointNetDeepLearning2017,meschederOccupancyNetworksLearning2019,boulchPOCOPointConvolution2022,zhang3DShape2VecSet3DShape2023} or transfer point data to voxel grids for further processing \citep{pengConvolutionalOccupancyNetworks2020,chibaneImplicitFunctionsFeature2020}.
To get the best of both we instead interleave these approaches.
The intuition behind this idea is, that the points represent exactly the shape that should be encoded.
In order to reason about the surrounding space of the object, the features are projected onto a grid and propagated by using shallow convolutional layers.
Then, to extract as much information as possible about the shape, features get interpolated from the grid to the input points and propagated between the nodes.
The usage of shallow 3D convolutions makes the network faster to evaluate and use less intermediate storage.

First, a graph convolution operator (e.g., EdgeConv by \citet{wangDynamicGraphCNN2019}) transforms each vertex $V^{f}$ using the edge connectivity information.
Next, we project the feature vectors onto a grid.
We do this using either element-wise max pooling from vertices located in that grid cell or our projection method described in Section \ref{sec:pic}.
A $2\times2\times2$ convolution and subsequent deconvolution with activation (where the feature count is doubled for the latent space to retain information) is then used to exchange information between neighboring cells.
We map the features back onto the vertices through tri-linear interpolation using the 8 closest cell centers.
Here, they are combined with the original output of the graph convolution before finally being processed through a single per-vertex fully-connected layer.
This output serves as the input of the next convolution block, while the deconvolved grid values are cached for later use.

We reason that the grid is suitable for distances in Euclidean space, while the graph convolution can maximize information extraction from the input points.
Combining both multiple times throughout the network makes the feature extraction time and memory efficient.

\paragraph{Encoder} %
The overall encoder-decoder architecture is shown in Figure~\ref{fig:architecture}.
At the beginning of the encoder, the input vertex positions $\mathcal{V}$ are transformed to per-vertex features $V^f$ through positional encoding \citep{tancikFourierFeaturesLet2020} and a single linear layer with ReLU activations.
We feed the encoded positions through a fixed number (typically 4) of our hybrid convolution blocks (as described above) with increasing resolution (contrary to the usual decreasing order of convolutional neural networks). 
The concatenated outputs of the grids form the grid vector (the feature vertex output of the last convolutional block can be discarded).
Values from each level of the grid vector are extracted for each of the $s$ SDF sample positions by tri-linear interpolation.
The results are summed element-wise to form the final latent vector.
We note that when latent vectors for additional sample positions are computed, there is no need to recompute the grid vector --- it can be cached for future use.

\paragraph{Decoder} The decoder receives the sample features and positions as input and processes them individually to compute a single signed distance value for each sample.
As architecture we make use of the proposed decoder of \citet{mehtaModulatedPeriodicActivations2021}, which is an augmentation of the SIREN layer proposed by \citet{sitzmannImplicitNeuralRepresentations2020}.
The feature vector is passed through a ReLU-activated MLP and the sample position through a Sine-activated MLP.
The activations of the ReLU MLP are then used to modulate the activations of the Sine MLP whose output is the final SDF value.

The decoder is specifically chosen to be a smaller network, since per our motivation, it should be efficient to evaluate for a very large number of query points.

\begin{table*}[tb]
   \caption{Metadata about each of our datasets. $\mathcal{O}$ denotes a roughly even split of manifold and non-manifold instances.}
   \label{tab:dataset_info}
   \begin{center}
      \begin{tabular}{lcccccccc}
         \toprule
         & Shapes Train & Shapes Test & Vertices Train & Vertices Test & Manifold &  Noisy & Sparse \\ 
         \cmidrule(lr){2-3}\cmidrule(lr){4-5}\cmidrule(lr){6-8}
         Dragon        & 2,400  & 300  & 2,210 & 2,210 & \cmark & \xmark & \xmark \\
         Armadillo     & 2,400  & 300  & 25,441 & 25,441 & \cmark & \xmark & \xmark \\
         DFaust (scans)& 6,258  & 2,038 & 30,000 & 100,000 & \xmark & \cmark & \xmark \\
         Thingi10k     & 2,000  & 200  & 4 - 4,995 & 282 - 4,890 & \cmark & \xmark & \cmark  \\
         ShapeNet V2 Planes     & 1,632  & 421  & 434 - 14,879 & 457 - 13,888 & \xmark & \xmark & \xmark \\
         Objaverse     & 2,000  & 200  & 50,000 & 100,000 & $\mathcal{O}$ & \xmark & \xmark \\
         \bottomrule
      \end{tabular}
   \end{center}
\end{table*}

\subsection{Loss Function}
\label{sec:loss}
An SDF can be defined as the unique solution $\Phi$ of the following eikonal equation:
\begin{equation}
\label{eq:eikonal_equation}
	\begin{aligned}
	\left\Vert\nabla \Phi(\vx)\right\Vert &= 1 \textnormal{ for } \vx \in \Omega\setminus\Omega_S \subset \R^3 \\
	\Phi(\vx) &= 0 \textnormal{ for } \vx \in \Omega_S,
	\end{aligned}
\end{equation}
where $\Omega$ is the SDF domain and $\Omega_S$ is the surface of the object.
Related to the work of \citet{smithEikoNetSolvingEikonal2021}, \citet{sitzmannImplicitNeuralRepresentations2020} proposed the following loss function as a measure of how well the eikonal equation is satisfied
\begin{equation}
\label{eq:eikonal_loss}
	\begin{aligned}
\gL_{\textnormal{eikonal}} = &\int_{\Omega} \left| \left\Vert\nabla\Phi(\vx)\right\Vert - 1  \right| \,\mathrm{d}\vx + \int_{\Omega_S} |\Phi(\vx)| \,\mathrm{d}\vx \\ 
&+ \int_{\Omega_S} \left(1 - \langle\nabla\Phi(\vx), \vn_x\rangle\right) \,\mathrm{d}\vx \\
&+ \int_{\Omega\setminus\Omega_S} \exp(-\alpha |\Phi(\vx)|) \,\mathrm{d}\vx.
	\end{aligned}
\end{equation}
Here $\Phi(\vx)$ denotes the predicted signed distance value of our neural network at position $\vx \in \Omega$, and $\vn_x$ denotes the target surface normal.
The exponential in the last term is a weighting function which ``pushes'' signed distance values away from zero for points not on the surface.
The original paper notes that the value of the constant $\alpha$ should be chosen as $\alpha \gg 1$, however we have found that this is sometimes detrimental to the desired effect, while choosing a lower value of $\alpha \approx 10$ yielded the best results.

If the input surface is non-manifold or self-intersecting, the inside and outside of the object's volume are not well defined and normal vectors do not necessarily point in the right direction.
For these cases we introduce a simple modification that ignores the sign of the normal:
\begin{equation}
\label{eq:normal_fix}
	\gL_{\textnormal{surface normal}} = \int_{\Omega_S} (1 - |\langle\nabla\Phi(\vx), \vn_x\rangle|) \,\mathrm{d}\vx .
\end{equation}
This change alone can still result in \emph{signed} distance fields.
We further propose the following change to train \emph{unsigned} distance fields, which can lead to better fitting of the geometry for the cases where an SDF cannot be recovered.
\begin{equation}
\label{eq:unsigned_fix}
	\gL_{\textnormal{non surface}} = \int_{\Omega\setminus\Omega_S} \exp(-\alpha \Phi(\vx)) \,\mathrm{d}\vx .
\end{equation}
As we discuss in Section \ref{sec:comparison}, this greatly improves performance on meshes with poor quality. 

\subsection{Point to Grid Transfer}
\label{sec:pic}
Instead of the usual voxelization schemes applied in neural networks, typically occupancy indicators, maximum pooling or average pooling, we propose to use the particle-in-cell interpolation approach \cite{harlowParticleincellMethodNumerical1962,jiangAffineParticleincellMethod2015}.
This approach is commonly used in hybrid physical simulation methods in order to transfer momentum from particles to an enclosing grid.
The concept of this transfer is sketched in Figure \ref{fig:grid-block} using grey arrows.
Assuming particle positions $\vx_p \in \R^3$, grid positions $\vx_i \in \R^3$ and unit masses $m_p = 1$, the feature vector $\vf_i$ at each grid point can be computed from the feature vectors at particle positions $\vf_p$ using
\begin{equation*}
	\hat{\vf}_i = \sum_p w_{ip}\vf_p, \quad w_i = \sum_p w_{ip}, \quad \vf_i = \frac{\hat{\vf}_i}{w_i}.
\end{equation*}
Here 
\begin{equation*}
	w_{ip}=N\left(\frac{x_p - x_i}{h}\right)N\left(\frac{y_p - y_i}{h}\right)N\left(\frac{z_p - z_i}{h}\right)
\end{equation*}
denotes the interpolation weights, $h$ the grid spacing and $N$ the linear interpolation weights given by
\begin{equation*}
	N(x) = \begin{cases}
		1 - |x|, & 0 \leq |x| < 1\\
		0, & 1\leq |x|.
	\end{cases}
\end{equation*}
Using this interpolation scheme instead of a pooling approach allows for a smoother transition between graph-convolutions and grid-convolutions, since both of them now approximate the same continuous ``feature-field''.
We show in Section \ref{sec:normals} that this improves reconstruction quality and reduces noise in reconstructed surfaces, particularly when used in conjunction with input normals.

\subsection{Training}
\label{sec:training}

A major advantage of our proposed architecture is that ground truth SDF values for the input point-cloud never have to be computed.
In order to evaluate the eikonal loss, only the sample positions and classification into (1) surface and (2) non-surface samples are required.

We sample them in the following ways: (1) Surface points. If the source is a triangle mesh rather than a point-cloud, its surface is randomly sampled to create additional on-surface samples. 
(2a) Random points in the surrounding volume are sampled as off-surface samples (no check is performed if these points actually lie on the surface per chance since the surface has zero measure as a volume). 
(2b) If surface normal information is available, additional close-to-surface samples are generated from random points on the surface by displacing them along the surface normal. 
These additional samples are not strictly needed but aid the training process, as the SDF is most detailed close to the surface.

We train our encoder-decoder architecture end to end by inputting batches of surface point-clouds and sample points and computing output signed distance values.
Automatic differentiation is used to provide spatial derivatives for the SDF gradients in $\gL_{\textnormal{eikonal}}$.

%% file: figures/architecture.pdf_tex
%% Creator: Inkscape 1.3 (0e150ed, 2023-07-21), www.inkscape.org
%% Accompanies image file 'architecture.pdf' (pdf, eps, ps)
\begingroup%
  \makeatletter%
  \providecommand\color[2][]{%
    \errmessage{(Inkscape) Color is used for the text in Inkscape, but the package 'color.sty' is not loaded}%
    \renewcommand\color[2][]{}%
  }%
  \providecommand\transparent[1]{%
    \errmessage{(Inkscape) Transparency is used (non-zero) for the text in Inkscape, but the package 'transparent.sty' is not loaded}%
    \renewcommand\transparent[1]{}%
  }%
  \providecommand\rotatebox[2]{#2}%
  \newcommand*\fsize{\dimexpr\f@size pt\relax}%
  \newcommand*\lineheight[1]{\fontsize{\fsize}{#1\fsize}\selectfont}%
  \ifx\svgwidth\undefined%
    \setlength{\unitlength}{510.29501343bp}%
    \ifx\svgscale\undefined%
      \relax%
    \else%
      \setlength{\unitlength}{\unitlength * \real{\svgscale}}%
    \fi%
  \else%
    \setlength{\unitlength}{\svgwidth}%
  \fi%
  \global\let\svgwidth\undefined%
  \global\let\svgscale\undefined%
  \makeatother%
  \begin{picture}(1,0.32656334)%
    \lineheight{1}%
    \setlength\tabcolsep{0pt}%
    \put(0.05250489,0.31748862){\makebox(0,0)[t]{\lineheight{1.25}\smash{\begin{tabular}[t]{c}Input point cloud\end{tabular}}}}%
    \put(0,0){\includegraphics[width=\unitlength,page=1]{architecture.pdf}}%
    \put(0.43770898,0.31599821){\color[rgb]{0,0,0}\makebox(0,0)[lt]{\lineheight{1.25}\smash{\begin{tabular}[t]{l}Interpolations\end{tabular}}}}%
    \put(0,0){\includegraphics[width=\unitlength,page=2]{architecture.pdf}}%
    \put(0.52674732,0.05783396){\color[rgb]{0,0,0}\makebox(0,0)[t]{\lineheight{1.25}\smash{\begin{tabular}[t]{c}(s, f)\end{tabular}}}}%
    \put(0,0){\includegraphics[width=\unitlength,page=3]{architecture.pdf}}%
    \put(0.63856085,0.13314031){\color[rgb]{0,0,0}\makebox(0,0)[t]{\lineheight{1.25}\smash{\begin{tabular}[t]{c}(s, 3)\end{tabular}}}}%
    \put(0,0){\includegraphics[width=\unitlength,page=4]{architecture.pdf}}%
    \put(0.85693778,0.17624001){\color[rgb]{0,0,0}\makebox(0,0)[t]{\lineheight{1.25}\smash{\begin{tabular}[t]{c}(s, 1)\end{tabular}}}}%
    \put(0.93146719,0.23981001){\color[rgb]{0,0,0}\makebox(0,0)[lt]{\lineheight{1.25}\smash{\begin{tabular}[t]{l}SDF\end{tabular}}}}%
    \put(0.20866408,0.31748862){\color[rgb]{0,0,0}\makebox(0,0)[t]{\lineheight{1.25}\smash{\begin{tabular}[t]{c}Convolution blocks\end{tabular}}}}%
    \put(0.31145663,0.12253507){\color[rgb]{0,0,0}\makebox(0,0)[t]{\lineheight{1.25}\smash{\begin{tabular}[t]{c}(4, 4, f)\end{tabular}}}}%
    \put(0,0){\includegraphics[width=\unitlength,page=5]{architecture.pdf}}%
    \put(0.1951238,0.08957001){\color[rgb]{0,0,0}\makebox(0,0)[t]{\lineheight{1.25}\smash{\begin{tabular}[t]{c}(n, f)\end{tabular}}}}%
    \put(0,0){\includegraphics[width=\unitlength,page=6]{architecture.pdf}}%
    \put(0.0887094,0.05429363){\color[rgb]{0,0,0}\makebox(0,0)[t]{\lineheight{1.25}\smash{\begin{tabular}[t]{c}(n, f)\end{tabular}}}}%
    \put(0,0){\includegraphics[width=\unitlength,page=7]{architecture.pdf}}%
    \put(0.20982098,0.16740297){\color[rgb]{0,0,0}\makebox(0,0)[t]{\lineheight{1.25}\smash{\begin{tabular}[t]{c}(n, f)\end{tabular}}}}%
    \put(0.22157873,0.2459965){\color[rgb]{0,0,0}\makebox(0,0)[t]{\lineheight{1.25}\smash{\begin{tabular}[t]{c}(n, f)\end{tabular}}}}%
    \put(0,0){\includegraphics[width=\unitlength,page=8]{architecture.pdf}}%
    \put(0.52674732,0.13485974){\color[rgb]{0,0,0}\makebox(0,0)[t]{\lineheight{1.25}\smash{\begin{tabular}[t]{c}(s, f)\end{tabular}}}}%
    \put(0,0){\includegraphics[width=\unitlength,page=9]{architecture.pdf}}%
    \put(0.29752001,0.04325377){\color[rgb]{0,0,0}\makebox(0,0)[t]{\lineheight{1.25}\smash{\begin{tabular}[t]{c}(2, 2, f)\end{tabular}}}}%
    \put(0.3234679,0.20121174){\color[rgb]{0,0,0}\makebox(0,0)[t]{\lineheight{1.25}\smash{\begin{tabular}[t]{c}(8, 8, f)\end{tabular}}}}%
    \put(0.33943268,0.27856237){\color[rgb]{0,0,0}\makebox(0,0)[t]{\lineheight{1.25}\smash{\begin{tabular}[t]{c}(16, 16, f)\end{tabular}}}}%
    \put(0,0){\includegraphics[width=\unitlength,page=10]{architecture.pdf}}%
    \put(0.52674732,0.21199724){\color[rgb]{0,0,0}\makebox(0,0)[t]{\lineheight{1.25}\smash{\begin{tabular}[t]{c}(s, f)\end{tabular}}}}%
    \put(0,0){\includegraphics[width=\unitlength,page=11]{architecture.pdf}}%
    \put(0.52674732,0.29043096){\color[rgb]{0,0,0}\makebox(0,0)[t]{\lineheight{1.25}\smash{\begin{tabular}[t]{c}(s, f)\end{tabular}}}}%
    \put(0.64443989,0.21995031){\color[rgb]{0,0,0}\makebox(0,0)[t]{\lineheight{1.25}\smash{\begin{tabular}[t]{c}(s, f)\end{tabular}}}}%
    \put(0,0){\includegraphics[width=\unitlength,page=12]{architecture.pdf}}%
    \put(0.3438334,-0.00000002){\color[rgb]{0,0,0}\makebox(0,0)[t]{\lineheight{1.25}\smash{\begin{tabular}[t]{c}Encoder\end{tabular}}}}%
    \put(0,0){\includegraphics[width=\unitlength,page=13]{architecture.pdf}}%
    \put(0.39884941,0.3159766){\color[rgb]{0,0,0}\makebox(0,0)[t]{\lineheight{1.25}\smash{\begin{tabular}[t]{c}Grid vector\\(340, f)\end{tabular}}}}%
    \put(0,0){\includegraphics[width=\unitlength,page=14]{architecture.pdf}}%
    \put(0.05787774,0.14549495){\color[rgb]{0,0,0}\makebox(0,0)[t]{\lineheight{1.25}\smash{\begin{tabular}[t]{c}Positional encoding\\+ FCL\end{tabular}}}}%
    \put(0,0){\includegraphics[width=\unitlength,page=15]{architecture.pdf}}%
    \put(0.07339482,0.17471172){\color[rgb]{0,0,0}\makebox(0,0)[t]{\lineheight{1.25}\smash{\begin{tabular}[t]{c}(n, 3)\end{tabular}}}}%
    \put(0,0){\includegraphics[width=\unitlength,page=16]{architecture.pdf}}%
    \put(0.74948031,-0.00000002){\color[rgb]{0,0,0}\makebox(0,0)[t]{\lineheight{1.25}\smash{\begin{tabular}[t]{c}Decoder\end{tabular}}}}%
    \put(0.61025259,0.18161787){\color[rgb]{0,0,0}\makebox(0,0)[t]{\lineheight{1.25}\smash{\begin{tabular}[t]{c}Query\\positions\end{tabular}}}}%
    \put(0,0){\includegraphics[width=\unitlength,page=17]{architecture.pdf}}%
    \put(0.73668956,0.25042728){\color[rgb]{0,0,0}\makebox(0,0)[t]{\lineheight{1.25}\smash{\begin{tabular}[t]{c}Modulated Siren MLP\end{tabular}}}}%
    \put(0,0){\includegraphics[width=\unitlength,page=18]{architecture.pdf}}%
    \put(0.29367325,0.26917483){\color[rgb]{0,0,0}\makebox(0,0)[t]{\lineheight{1.25}\smash{\begin{tabular}[t]{c}\textbf{G}\end{tabular}}}}%
    \put(0.29367325,0.29032297){\color[rgb]{0,0,0}\makebox(0,0)[t]{\lineheight{1.25}\smash{\begin{tabular}[t]{c}\textbf{F}\end{tabular}}}}%
    \put(0,0){\includegraphics[width=\unitlength,page=19]{architecture.pdf}}%
    \put(0.2401738,0.27857816){\color[rgb]{0,0,0}\makebox(0,0)[t]{\lineheight{1.25}\smash{\begin{tabular}[t]{c}16\end{tabular}}}}%
    \put(0,0){\includegraphics[width=\unitlength,page=20]{architecture.pdf}}%
    \put(0.22129053,0.20034057){\color[rgb]{0,0,0}\makebox(0,0)[lt]{\lineheight{1.25}\smash{\begin{tabular}[t]{l}8\end{tabular}}}}%
    \put(0.27897587,0.19119194){\color[rgb]{0,0,0}\makebox(0,0)[t]{\lineheight{1.25}\smash{\begin{tabular}[t]{c}\textbf{G}\end{tabular}}}}%
    \put(0.27897587,0.21234009){\color[rgb]{0,0,0}\makebox(0,0)[t]{\lineheight{1.25}\smash{\begin{tabular}[t]{c}\textbf{F}\end{tabular}}}}%
    \put(0,0){\includegraphics[width=\unitlength,page=21]{architecture.pdf}}%
    \put(0.20953262,0.12210299){\color[rgb]{0,0,0}\makebox(0,0)[lt]{\lineheight{1.25}\smash{\begin{tabular}[t]{l}4\end{tabular}}}}%
    \put(0.26721797,0.11267124){\color[rgb]{0,0,0}\makebox(0,0)[t]{\lineheight{1.25}\smash{\begin{tabular}[t]{c}\textbf{G}\end{tabular}}}}%
    \put(0.26721797,0.13381939){\color[rgb]{0,0,0}\makebox(0,0)[t]{\lineheight{1.25}\smash{\begin{tabular}[t]{c}\textbf{F}\end{tabular}}}}%
    \put(0,0){\includegraphics[width=\unitlength,page=22]{architecture.pdf}}%
    \put(0.19483524,0.04386539){\color[rgb]{0,0,0}\makebox(0,0)[lt]{\lineheight{1.25}\smash{\begin{tabular}[t]{l}2\end{tabular}}}}%
    \put(0.25252058,0.03486761){\color[rgb]{0,0,0}\makebox(0,0)[t]{\lineheight{1.25}\smash{\begin{tabular}[t]{c}\textbf{G}\end{tabular}}}}%
    \put(0.25252058,0.05601576){\color[rgb]{0,0,0}\makebox(0,0)[t]{\lineheight{1.25}\smash{\begin{tabular}[t]{c}\textbf{F}\end{tabular}}}}%
    \put(0,0){\includegraphics[width=\unitlength,page=23]{architecture.pdf}}%
  \end{picture}%
\endgroup%

%% file: sections/4-results.tex
\section{Results}
\label{sec:results}

% Ablation table
\input{data/scripts/ablation.tex}

The evaluation of our method is split into four parts:
First, we perform an ablation study to motivate different design decisions in our architecture. 
Secondly, we compare different model sizes (i.e., number of learned parameters) with respect to the number of details they can reconstruct. 
Thirdly, we compare our results to a number of recent encoder architectures to highlight the advantages of our approach.
Finally, we showcase the improvement in our method when using the improved point-to-grid feature projection, which is reinforced when using surface point normals as additional input.

\paragraph{Datasets}
To cover a variety of relevant scenarios we use six different datasets, which are summarized in Table \ref{tab:dataset_info}.
The first two datasets consist of synthetic deformations of two different simulated solid objects, a low-resolution dragon and a high-resolution armadillo, which were computed using the IPC simulation framework \citep{liIncrementalPotentialContact2020}. 
We additionally use the raw point scan data of the dynamic FAUST (DFaust) dataset \citep{bogoDynamicFAUSTRegistering2017}, which consists of temporally deforming 3D shapes with holes and noise.
To analyze performance on variable vertex count as well as single-class and multi-class encoding, we employ the Thingi10k dataset as provided by \citet{huTetrahedralMeshingWild2018} as well as the ``planes'' category of the ShapeNetV2 dataset \citep{changShapeNetInformationrich3D2015}.
Finally, we use the public-domain models of the Objaverse dataset \citep{deitkeObjaverseUniverseAnnotated2023} to test the scalability of our method.

\paragraph{Baselines.}
We focus our comparison on other encoder-based 3D shape representation methods, as they are most similar to our work.
To that end we implemented the encoder components of Occupancy Networks (ONet) \citep{meschederOccupancyNetworksLearning2019}, Convolutional Occupancy Networks (ConvONet) \citep{pengConvolutionalOccupancyNetworks2020}, Implicit Feature Networks (IFNet) \citep{chibaneImplicitFunctionsFeature2020}, Point Convolution for Surface Reconstruction (POCO) \citep{boulchPOCOPointConvolution2022} and 3DShape2VecSet (3DS2VS) \citep{zhang3DShape2VecSet3DShape2023}.
To ensure a fair comparison, we use the same training and testing data for all models as well as the same modulated-SIREN decoder as discussed in Section~\ref{sec:method}. 
The implementations of POCO and 3DS2VS were unfortunately not trivially adjustable to account for a variable number of input points during training, which is why they are not evaluated for our versions of the Thingi10K and ShapeNet dataset.
For all comparisons with the other methods, our model utilizes five layers of latent grids with resolutions [4,8,16,32,64] and a latent size of 64.
Furthermore, all beneficial options discussed in the ablation study in Section \ref{sec:ablation} are enabled, except for artificial noise.
This is omitted to ensure fair comparisons, as different models may react differently to equal levels of added noise.

\paragraph{Metrics.}
We focus our evaluation on two different metrics:
The widely used Chamfer distance (CD) and normal consistency (NC). The Chamfer distance compares distances between two point clouds $A$ and $B$ by finding the distance to the closest point in $B$ for each point in $A$ and taking the average.
The normal consistency instead compares orientations of normals by also finding the closest point in $B$ for each point in $A$, but instead computing the cosine similarity of the normal vectors and taking the average.
The same is done for the reverse direction and the results are summed together in the case of the Chamfer distance, and averaged in the case of the normal consistency.
For all tests we use 200-2000 individual reconstructions and report mean values and standard deviations for both metrics.

\paragraph{Hardware and Framework.} 
We train all models on either RTX 2080 Ti GPUs with 12GB of memory or RTX 4000 Ada GPUs with 20GB of memory.
With this, the training time for 400 epochs of our model using a batch size of 16 on a single GPU is between 1-2 days.
Most other models trained in similar time, apart from IFNet, POCO, and 3DS2VS which required between 3-16 days on a single GPU.
We implemented all of the models in this paper in the PyTorch framework \citep{paszkePyTorchImperativeStyle2019}.
In addition, we make use of PyTorch-Geometric for its 3D graph and set learning capabilities \citep{feyFastGraphRepresentation2019}.
All models were trained using the Adam optimizer using a learning rate of 5e-4 and a batch size of 16.
We have also implemented the tri-linear interpolation and particle-in-cell interpolation using custom CUDA kernels.

\subsection{Ablation}
\label{sec:ablation}

To highlight the impact of different design choices within our network architecture, we first conduct an ablation study covering three decisions within our convolution block in Figure~\ref{fig:grid-block}:
1) using the nearest neighbor instead of linear interpolation to map values from latent grid back to the input points,
2) enabling/ disabling the graph convolutions (GNN), and
3) enabling/ disabling the grid convolutions (CNN).

The results for different combinations are reported in Table~\ref{tab:ablation}.
The most impactful component is clearly using the grid convolutions, which also contains the majority of trainable weights.
The next important decision is enabling the graph convolution.
Interestingly, the impact of enabling graph convolutions is greater when interpolation is also enabled.
It also seems, that interpolation only has an impact on network performance when the graph convolution is also enabled.
This is because without the graph convolution, the interpolated features are immediately pooled back to the grid, eliminating most implicit information gain.
For maximal performance, all options should enabled.

% Knn Ablation
\input{scripts/knn_ablation.tex}
% Noise ablation table
\input{scripts/noise_ablation.tex}

Next, we investigate the impact of varying the number of k-NN for connecting the input point cloud, using point convolutions \cite{qiPointNetDeepLearning2017} instead of edge convolutions \cite{wangDynamicGraphCNN2019} and finally reversing the resolution of latent grids.
We show the results in Table \ref{tab:knn_ablation}.
In the depicted parameter sequence, the grid order has the single largest impact.
This suggests that coarse features are extracted earlier and fine features later in the network.
Using an increasing number of k-NN consistently improves both CD and NC in terms of average and spread.
For all experiments we use the settings corresponding to the best result.

Finally, we investigate our architectures sensitivity to noise by training and testing using different noise levels.
We add noise by perturbing input vertices along the normal direction by a random amount drawn from $\mathcal{U}(-\sigma,\sigma)$, where $\sigma$ is the noise level shown in the column labels and row labels in Table \ref{tab:noise_ablation}.
Since the coordinates have been normalized to the unit box, the maximum evaluated noise of 5e-2 corresponds to a shift of $1/40$ of the bounding box edge length. 
Our model remains stable during training up to noise levels of 1e-2 and even benefits from data augmentation using noise.

\subsection{Model Size}
\begin{figure}[tb]
   \def\svgwidth{\linewidth}
   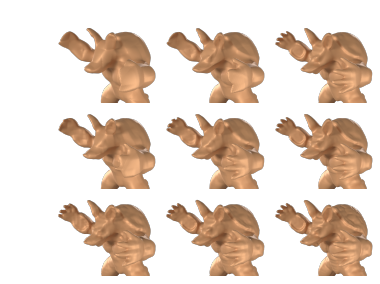
   \caption{
From top to bottom we increase the size of all latent grids, while from left to right the size of the latent feature is increased. 
Below each figure the inference time for 17M points ($256^3$ regular grid) and the storage requirements for the latent grid is shown.
The number of trainable network parameters for feature sizes 16, 32, and 64 are 38K, 151K, and 604K, respectively, regardless of the specific grid sizes.
}
   \label{fig:grid_res_comparison}
\end{figure}
The number of features per vertex and the resolutions of the grids are tunable parameters of our architecture.
Increasing them enables the network to capture more surface details, at the cost of increased storage requirements or computation time.
The results are shown in Figure~\ref{fig:grid_res_comparison}.
We find that the network behaves very intuitively and results degraded gracefully when the model size is reduced.
Most notably, the models in the first column with a latent size of 16 all contain only 38K network parameters in total, of which the decoder contains just 1.4K.
High-frequency features vanish first which makes small model sizes particularly appealing for applications such as approximate collision tests, e.g., in physical simulation, or deployment in resource constrained environments, e.g., edge devices.

\subsection{Comparison to Related Work}
\label{sec:comparison}

\begin{figure*}[tb]
   \centering
   \fontsize{8}{9}\selectfont
   \def\svgwidth{.95\linewidth}
   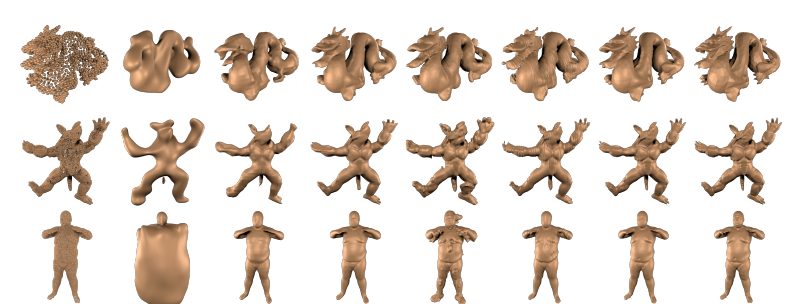\\
   (a) Fixed vertex-count datasets. \\\vspace*{1em}
   \def\svgwidth{.95\linewidth}
   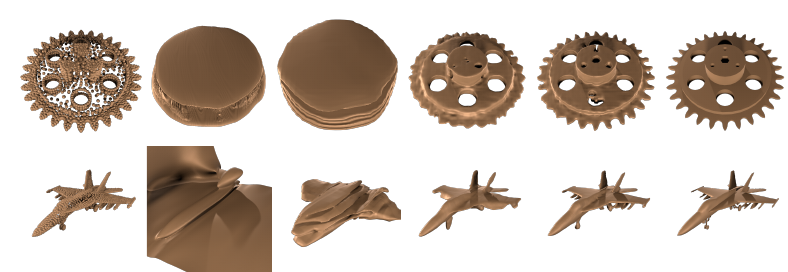\\
   (b) Variable vertex-count datasets. \\\vspace*{1em}
   \caption{Comparing our method to related baselines. Please refer to the accompanying video for a more immersive comparison. }
   \label{fig:main_comparison}
\end{figure*}
\begin{table*}[tb]
\caption{A summary of the comparisons between our encoder method, ONet \citep{meschederOccupancyNetworksLearning2019}, ConvONet \citep{pengConvolutionalOccupancyNetworks2020}, IFNet \citep{chibaneImplicitFunctionsFeature2020}, POCO \citep{boulchPOCOPointConvolution2022} and 3DShape2VecSet \citep{zhang3DShape2VecSet3DShape2023}. The table shows the Chamfer distance (CD) and normal consistency (NC). The arrow indicates whether lower or higher values are better. The best score is marked using a bold font.} 
\label{tab:main_results}
\begin{center}
\fontsize{8}{9}\selectfont
\input{data/scripts/main_results.tex}%
Parameters & \multicolumn{2}{c}{626K} & \multicolumn{2}{c}{1.1M}  & \multicolumn{2}{c}{1.9M} & \multicolumn{2}{c}{12.8M} & \multicolumn{2}{c}{106M} & \multicolumn{2}{c}{747K} \\
Inference & \multicolumn{2}{c}{2.45 s} & \multicolumn{2}{c}{0.7 s}  & \multicolumn{2}{c}{15.3 s}  & \multicolumn{2}{c}{120 s - 330 s} & \multicolumn{2}{c}{12.3 s - 20.1 s} & \multicolumn{2}{c}{2.34 s}  \\
\bottomrule
\end{tabular}
\end{center}
\end{table*}

We now discuss comparisons to the other baseline shape encoder methods on five datasets.
They are shown for all methods and datasets in the accompanying video, as well as in Figure \ref{fig:main_comparison} and Table \ref{tab:main_results}.
We can summarize that our method outperforms the other baselines with respect to reconstructed surface detail, both in terms of the Chamfer distance and normal consistency.
Only for the Dragon and Thingi10k dataset, was the normal consistency of IFNet within 2\% of our method.
However, for these datasets the visual comparison in Figure \ref{fig:main_comparison} shows that our method is able to capture more detailed surfaces overall, as also indicated by the smaller Chamfer distance.
Note especially the spine and claws of the dragon, and the distinct teeth on the cog-wheel.
The cog-wheel unfortunately also shows a potential drawback of using k-nearest neighbor connectivity, where fine detail might become ``connected''. 
This can be mitigated by using either point-to-grid projection, oriented point-clouds, or both, as shown in Section \ref{sec:normals}.
Nevertheless, our architecture is still able to capture fine details better than all other baselines.

Our method also is able to reconstruct a dense grid of approx.~17M points in only 2.35\,s, while IFNet takes more than 6x longer.
Only ConvONet is faster on the reconstruction task, however at significantly lower quality. 
We also find that our approach offers the highest quality per parameter.

Our model consistently exhibits very low standard deviation on all metrics, which makes the performance of the network more predictable.
This is underlined by the comparison in Figure \ref{fig:grid_res_comparison}, which shows the effect of changing latent size and grid resolutions.
Very small networks with our architecture can already have remarkable representational abilities.

Finally, we introduced a simple modification, see \Eqref{eq:normal_fix} and \Eqref{eq:unsigned_fix}, to the loss function to be able to compute distance fields for surfaces with inconsistent surface normals, which we tested on the ShapeNetV2 dataset.
We observed an improvement on all metrics for three of the four tested methods, with the exception of ONets \citep{meschederOccupancyNetworksLearning2019}, which partially performed better using the original loss.
Although it should be noted that the model did not seem to be able to deal well with the changing number of vertices in the dataset, which is why we interpret this result as an outlier.

\footnotetext[1]{Evaluated on a subset of 200 random shapes due to computational constraints.}

\subsection{Point-to-Grid Feature Projection}
\label{sec:normals}

\begin{table}[t]
   \caption{Comparing our method with pooling-based and projection-based point-to-grid transfers.
   Optionally, input vertex normals are also used. 
   We evaluate the Chamfer distance (CD) and normal consistency (NC).
   Bold font marks the best value in each metric and dataset.
   All values are given scaled to $\cdot 10 ^{-2}$.
   }
   \label{tab:normals}
   \centering
   \fontsize{8}{9}\selectfont

\input{data/scripts/normals_comparison.tex}
   \end{tabular}
\end{table}

\begin{figure}[t]
   \centering
   \fontsize{7}{9}\selectfont
   \def\svgwidth{\linewidth}
   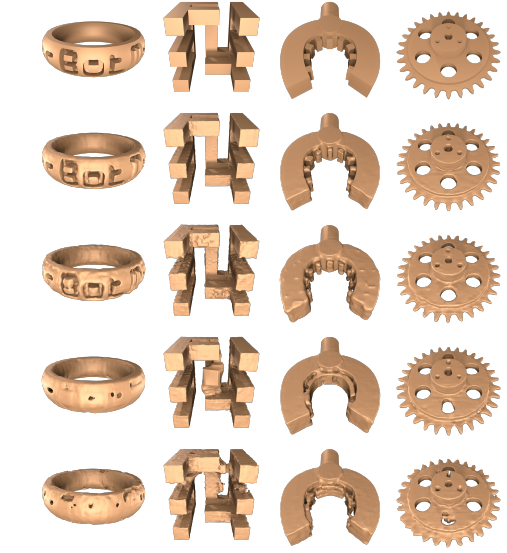
   \caption{Visual comparison of our architecture using pooling-based and projection-based point-to-grid transfers.
   Row two and three from the top use point normals as extra input.}
   \label{fig:normals_proj}
\end{figure}

As described in Section \ref{sec:pic}, we propose a physically-inspired method to project particle features to the convolutional grid.
Simply exchanging the pooling operator for this projection already results in improved metrics for most datasets, see the left two columns of Table \ref{tab:normals}. 
This exchange comes at identical network complexity and memory requirements and nearly identical training and inference speed.
The feature projection not only improves the metrics, but also makes reconstructions smoother and details sharper, as can be seen in Figure \ref{fig:normals_proj}.

In the case that the input point clouds contain information about orientation, i.e. point-normals, these can be used as additional inputs to our network.
To do so, they are concatenated to the surface-point positions (cf. Figure \ref{fig:architecture}). 
Results using normals are shown in the right two columns of Table \ref{tab:normals} for the pooling-based and projection-based variant of our network, respectively.
Again, projection measurably improves performance, outperforming all other variants of our model. 
The reconstructed surfaces in Figure \ref{fig:normals_proj} again show improved surface smoothness and detail sharpness. 
The reader is encouraged to view examples from the other datasets in the second half of the supplementary video.
Since projection maps a single point to eight grid cells in contrast the one to one mapping of pooling, we have at times found it useful to apply dropout within the projection operator to prevent overfitting.

%% file: data/scripts/ablation.tex
\begin{table}[tb]
\centering
\caption{The results of comparing different design choices within the structure of our network (see Figure~\ref{fig:grid-block}). The tables show the Chamfer distance (CD) and normal consistency (NC). The cell color coding is derived from the mean value. The arrow indicates whether lower or higher values are better. A darker color corresponds to a better value.}
\label{tab:ablation}
\fontsize{8}{9}\selectfont
% \begin{tabular}{cc}
\begin{subtable}[t]{\linewidth}
\centering
\caption{Chamfer Distance $\downarrow$ mean / std ($\cdot10^{-2}$)}
\label{tab:ablation_chamfer_loss}
\begin{tabular}{p{10mm}p{11mm}p{11mm}p{11mm}p{11mm}p{11mm}p{11mm}p{11mm}p{11mm}}
\toprule
 & \multicolumn{2}{c}{Interpolation on} & \multicolumn{2}{c}{Interpolation off}\\
\cmidrule(lr){2-3} \cmidrule(lr){4-5}
 GNN & CNN on & CNN off & CNN on & CNN off\\
\midrule
on  &  \cellcolor[RGB]{221.35294117647058, 41.843137254901954, 36.647058823529406} 2.626\,/ 0.257 &  \cellcolor[RGB]{252.0, 171.24313725490197, 142.93725490196078} 3.402\,/ 0.606 &  \cellcolor[RGB]{242.90588235294118, 74.29803921568626, 53.764705882352935} 2.834\,/ 0.367 &  \cellcolor[RGB]{253.74117647058824, 219.21176470588236, 203.65882352941176} 3.715\,/ 0.939 \\
off  &  \cellcolor[RGB]{248.92941176470586, 97.89019607843136, 68.8235294117647} 2.959\,/ 0.442 &  \cellcolor[RGB]{255.0, 245.0, 240.0} 4.035\,/ 1.036 &  \cellcolor[RGB]{248.92941176470586, 97.89019607843136, 68.8235294117647} 2.957\,/ 0.452 &  \cellcolor[RGB]{255.0, 245.0, 240.0} 4.038\,/ 1.091 \\
\bottomrule
\end{tabular}
\end{subtable}
\hfill \vspace{5pt}\\
\begin{subtable}[t]{\linewidth}
\centering
\caption{Normal Consistency $\uparrow$ mean / std ($\cdot10^{-2}$)}
\label{tab:ablation_normal_consistency}
\begin{tabular}{p{10mm}p{11mm}p{11mm}p{11mm}p{11mm}p{11mm}p{11mm}p{11mm}p{11mm}}
\toprule
 & \multicolumn{2}{c}{Interpolation on} & \multicolumn{2}{c}{Interpolation off}\\
\cmidrule(lr){2-3} \cmidrule(lr){4-5}
 GNN & CNN on & CNN off & CNN on & CNN off\\
\midrule
on  &  \cellcolor[RGB]{109.62352941176471, 88.2470588235294, 166.78823529411764} 95.64\,/ 1.390 &  \cellcolor[RGB]{215.8823529411765, 215.95294117647057, 233.94117647058823} 88.15\,/ 6.970 &  \cellcolor[RGB]{122.04705882352941, 113.09411764705882, 179.77647058823527} 94.54\,/ 2.280 &  \cellcolor[RGB]{217.76470588235293, 217.77254901960782, 234.88235294117644} 88.00\,/ 7.345 \\
off  &  \cellcolor[RGB]{136.8235294117647, 133.52941176470588, 190.1176470588235} 93.39\,/ 3.220 &  \cellcolor[RGB]{251.1843137254902, 250.12156862745096, 252.4980392156863} 84.17\,/ 9.005 &  \cellcolor[RGB]{136.8235294117647, 133.52941176470588, 190.1176470588235} 93.41\,/ 3.228 &  \cellcolor[RGB]{252.0, 251.0, 253.0} 84.00\,/ 9.122 \\
\bottomrule
\end{tabular}
\end{subtable}
% \end{tabular}
\end{table}

%% file: scripts/knn_ablation.tex
\begin{table}[t]
\centering
\caption{Ablation study of the influence of the grid order, graph convolution and the number of nearest neighbors on the performance of our method. CD denotes the Chamfer Distance and NC the Normal Consistency. The best values are highlighted in bold. All values denote the mean and standard deviation.}
\label{tab:knn_ablation}
\fontsize{8}{9}\selectfont
\begin{tabular}{cccccc}
\toprule
GNN & Grid Order & KNN & CD $\downarrow (\cdot 10^{-2})$&NC $\uparrow (\cdot 10^{-2})$\\
\midrule
PointConv & Decr. & None & $0.984 / 0.052$ & $97.677 / 0.731$ \\
PointConv & Incr. & None & $0.953 / 0.049$ & $97.929 / 0.586$ \\
EdgeConv & Incr. & 2 & $0.937 / 0.049$ & $98.031 / 0.584$ \\
EdgeConv & Incr. & 4 & $0.941 / 0.047$ & $98.036 / 0.571$ \\
EdgeConv & Incr. & 8 & $\mathbf{0.935 / 0.047}$ & $\mathbf{98.089 / 0.543}$ \\
\bottomrule
\end{tabular}
\end{table}

%% file: scripts/noise_ablation.tex
\begin{table}[tb]
\centering
\caption{Comparison of Chamfer Loss and Normal Consistency across different noise levels. Rows indicate the noise level used during training and columns during testing. The best value for each metric is marked in bold font.}
\label{tab:noise_ablation}
\fontsize{8}{9}\selectfont
% \begin{tabular}{cc}
\begin{subtable}[t]{\linewidth}
\centering
\caption{Chamfer Distance $\downarrow$ mean / std $(\cdot 10^{-2})$}
\label{tab:noise_ablation_chamfer_loss}
\begin{tabular}{lccccc}
\toprule
Train \textbackslash Test & none & 5e-4 & 5e-3 & 1e-2 & 5e-2 \\
\midrule
none & $3.421$ & $3.421$ & $3.433$ & $3.508$ & $4.400$ \\
5e-4 & $\mathbf{3.366}$ & $\mathbf{3.366}$ & $3.407$ & $3.521$ & $4.539$ \\
5e-3 & $3.400$ & $3.401$ & $3.434$ & $3.540$ & $4.579$ \\
1e-2 & $3.528$ & $3.524$ & $3.540$ & $3.634$ & $4.733$ \\
\bottomrule
\end{tabular}
\end{subtable}
\hfill\vspace{5pt}\\
\begin{subtable}[t]{\linewidth}
\centering
\caption{Normal Consistency $\uparrow$ mean / std $(\cdot 10^{-2})$}
\label{tab:noise_ablation_normal_consistency}
\begin{tabular}{lccccc}
\toprule
Train \textbackslash Test & none & 5e-4 & 5e-3 & 1e-2 & 5e-2 \\
\midrule
none & $92.860$ & $92.835$ & $92.642$ & $91.679$ & $75.103$ \\
5e-4 & $93.136$ & $93.146$ & $92.902$ & $91.774$ & $76.036$ \\
5e-3 & $93.172$ & $93.159$ & $93.073$ & $92.381$ & $77.292$ \\
1e-2 & $93.208$ & $93.222$ & $\mathbf{93.273}$ & $93.078$ & $80.211$ \\
\bottomrule
\end{tabular}
\end{subtable}
% \end{tabular}
\end{table}

%% file: figures/comparison/grid_resolution.pdf_tex
%% Creator: Inkscape 1.3.2 (091e20e, 2023-11-25), www.inkscape.org
%% Accompanies image file 'grid_resolution.pdf' (pdf, eps, ps)
\begingroup%
  \makeatletter%
  \providecommand\color[2][]{%
    \errmessage{(Inkscape) Color is used for the text in Inkscape, but the package 'color.sty' is not loaded}%
    \renewcommand\color[2][]{}%
  }%
  \providecommand\transparent[1]{%
    \errmessage{(Inkscape) Transparency is used (non-zero) for the text in Inkscape, but the package 'transparent.sty' is not loaded}%
    \renewcommand\transparent[1]{}%
  }%
  \providecommand\rotatebox[2]{#2}%
  \newcommand*\fsize{\dimexpr\f@size pt\relax}%
  \newcommand*\lineheight[1]{\fontsize{\fsize}{#1\fsize}\selectfont}%
  \ifx\svgwidth\undefined%
    \setlength{\unitlength}{182.36061859bp}%
    \ifx\svgscale\undefined%
      \relax%
    \else%
      \setlength{\unitlength}{\unitlength * \real{\svgscale}}%
    \fi%
  \else%
    \setlength{\unitlength}{\svgwidth}%
  \fi%
  \global\let\svgwidth\undefined%
  \global\let\svgscale\undefined%
  \makeatother%
  \begin{picture}(1,0.74719669)%
    \lineheight{1}%
    \setlength\tabcolsep{0pt}%
    \put(0,0){\includegraphics[width=\unitlength,page=1]{grid_resolution.pdf}}%
    \put(0.57583745,0.72616447){\makebox(0,0)[t]{\lineheight{1.25}\smash{\begin{tabular}[t]{c}\small Increasing feature size\end{tabular}}}}%
    \put(0.03704131,0.35212159){\rotatebox{90}{\makebox(0,0)[t]{\lineheight{1.25}\smash{\begin{tabular}[t]{c}\small Increasing grid size\end{tabular}}}}}%
    \put(0.28897683,0.67893096){\makebox(0,0)[t]{\lineheight{1.25}\smash{\begin{tabular}[t]{c}\small 16\end{tabular}}}}%
    \put(0.57527015,0.67879259){\makebox(0,0)[t]{\lineheight{1.25}\smash{\begin{tabular}[t]{c}\small 32\end{tabular}}}}%
    \put(0.86136977,0.67887561){\makebox(0,0)[t]{\lineheight{1.25}\smash{\begin{tabular}[t]{c}\small 64\end{tabular}}}}%
    \put(0,0){\includegraphics[width=\unitlength,page=2]{grid_resolution.pdf}}%
    \put(0.09103656,0.56916639){\makebox(0,0)[t]{\smash{\begin{tabular}[t]{c}\small 1x\end{tabular}}}}%
    \put(0.09171458,0.34208977){\makebox(0,0)[t]{\smash{\begin{tabular}[t]{c}\small 2x\end{tabular}}}}%
    \put(0.09113342,0.11501339){\makebox(0,0)[t]{\lineheight{1.25}\smash{\begin{tabular}[t]{c}\small 4x\end{tabular}}}}%
    \put(0.57574562,0.45835272){\makebox(0,0)[t]{\lineheight{1.25}\smash{\begin{tabular}[t]{c}\tiny 1.05 s / 0.60 MB\end{tabular}}}}%
    \put(0.29000958,0.45835272){\makebox(0,0)[t]{\lineheight{1.25}\smash{\begin{tabular}[t]{c}\tiny 0.56 s / 0.30 MB\end{tabular}}}}%
    \put(0.86105904,0.45836279){\makebox(0,0)[t]{\lineheight{1.25}\smash{\begin{tabular}[t]{c}\tiny 1.95 s / 1.2 MB\end{tabular}}}}%
    \put(0.29000958,0.23127614){\makebox(0,0)[t]{\lineheight{1.25}\smash{\begin{tabular}[t]{c}\tiny 0.58 s / 2.40 MB\end{tabular}}}}%
    \put(0.57601734,0.23127614){\makebox(0,0)[t]{\lineheight{1.25}\smash{\begin{tabular}[t]{c}\tiny 1.08 s / 4.80 MB\end{tabular}}}}%
    \put(0.86150187,0.23128621){\makebox(0,0)[t]{\lineheight{1.25}\smash{\begin{tabular}[t]{c}\tiny 1.99 s / 9.58 MB\end{tabular}}}}%
    \put(0.28960704,0.0042096){\makebox(0,0)[t]{\lineheight{1.25}\smash{\begin{tabular}[t]{c}\tiny 0.63 s / 19.17 MB\end{tabular}}}}%
    \put(0.57572554,0.0042096){\makebox(0,0)[t]{\lineheight{1.25}\smash{\begin{tabular}[t]{c}\tiny 1.19 s / 38.34 MB\end{tabular}}}}%
    \put(0.86151188,0.0042096){\makebox(0,0)[t]{\lineheight{1.25}\smash{\begin{tabular}[t]{c}\tiny 2.33 s / 76.67 MB\end{tabular}}}}%
  \end{picture}%
\endgroup%

%% file: figures/comparison/main_comparison_fixed_vertex.pdf_tex
%% Creator: Inkscape 1.3.2 (091e20e, 2023-11-25), www.inkscape.org
%% Accompanies image file 'main_comparison_fixed_vertex.pdf' (pdf, eps, ps)
\begingroup%
  \makeatletter%
  \providecommand\color[2][]{%
    \errmessage{(Inkscape) Color is used for the text in Inkscape, but the package 'color.sty' is not loaded}%
    \renewcommand\color[2][]{}%
  }%
  \providecommand\transparent[1]{%
    \errmessage{(Inkscape) Transparency is used (non-zero) for the text in Inkscape, but the package 'transparent.sty' is not loaded}%
    \renewcommand\transparent[1]{}%
  }%
  \providecommand\rotatebox[2]{#2}%
  \newcommand*\fsize{\dimexpr\f@size pt\relax}%
  \newcommand*\lineheight[1]{\fontsize{\fsize}{#1\fsize}\selectfont}%
  \ifx\svgwidth\undefined%
    \setlength{\unitlength}{379.35710907bp}%
    \ifx\svgscale\undefined%
      \relax%
    \else%
      \setlength{\unitlength}{\unitlength * \real{\svgscale}}%
    \fi%
  \else%
    \setlength{\unitlength}{\svgwidth}%
  \fi%
  \global\let\svgwidth\undefined%
  \global\let\svgscale\undefined%
  \makeatother%
  \begin{picture}(1,0.38459703)%
    \lineheight{1}%
    \setlength\tabcolsep{0pt}%
    \put(0.08358984,0.36856993){\color[rgb]{0,0,0}\makebox(0,0)[t]{\lineheight{1.25}\smash{\begin{tabular}[t]{c}Input\end{tabular}}}}%
    \put(0.20518007,0.36856993){\color[rgb]{0,0,0}\makebox(0,0)[t]{\lineheight{1.25}\smash{\begin{tabular}[t]{c}ONet\end{tabular}}}}%
    \put(0.32677029,0.36856993){\color[rgb]{0,0,0}\makebox(0,0)[t]{\lineheight{1.25}\smash{\begin{tabular}[t]{c}ConvONet\end{tabular}}}}%
    \put(0.4483605,0.36856993){\color[rgb]{0,0,0}\makebox(0,0)[t]{\lineheight{1.25}\smash{\begin{tabular}[t]{c}IFNet\end{tabular}}}}%
    \put(0.56995077,0.36856993){\color[rgb]{0,0,0}\makebox(0,0)[t]{\lineheight{1.25}\smash{\begin{tabular}[t]{c}POCO\end{tabular}}}}%
    \put(0.69154095,0.36856993){\color[rgb]{0,0,0}\makebox(0,0)[t]{\lineheight{1.25}\smash{\begin{tabular}[t]{c}3DS2VS\end{tabular}}}}%
    \put(0.81313119,0.36856993){\color[rgb]{0,0,0}\makebox(0,0)[t]{\lineheight{1.25}\smash{\begin{tabular}[t]{c}Ours\end{tabular}}}}%
    \put(0.93472136,0.36856993){\color[rgb]{0,0,0}\makebox(0,0)[t]{\lineheight{1.25}\smash{\begin{tabular}[t]{c}Ground Truth\end{tabular}}}}%
    \put(0.01602709,0.30100718){\color[rgb]{0,0,0}\rotatebox{90}{\makebox(0,0)[t]{\lineheight{0}\smash{\begin{tabular}[t]{c}Dragon\end{tabular}}}}}%
    \put(0.01602709,0.17941695){\color[rgb]{0,0,0}\rotatebox{90}{\makebox(0,0)[t]{\lineheight{0}\smash{\begin{tabular}[t]{c}Armadillo\end{tabular}}}}}%
    \put(0.01602709,0.05782673){\color[rgb]{0,0,0}\rotatebox{90}{\makebox(0,0)[t]{\lineheight{0}\smash{\begin{tabular}[t]{c}DFaust\end{tabular}}}}}%
    \put(0,0){\includegraphics[width=\unitlength,page=1]{main_comparison_fixed_vertex.pdf}}%
  \end{picture}%
\endgroup%

%% file: figures/comparison/main_comparison_variable_vertex.pdf_tex
%% Creator: Inkscape 1.3 (0e150ed, 2023-07-21), www.inkscape.org
%% Accompanies image file 'main_comparison_variable_vertex.pdf' (pdf, eps, ps)
\begingroup%
  \makeatletter%
  \providecommand\color[2][]{%
    \errmessage{(Inkscape) Color is used for the text in Inkscape, but the package 'color.sty' is not loaded}%
    \renewcommand\color[2][]{}%
  }%
  \providecommand\transparent[1]{%
    \errmessage{(Inkscape) Transparency is used (non-zero) for the text in Inkscape, but the package 'transparent.sty' is not loaded}%
    \renewcommand\transparent[1]{}%
  }%
  \providecommand\rotatebox[2]{#2}%
  \newcommand*\fsize{\dimexpr\f@size pt\relax}%
  \newcommand*\lineheight[1]{\fontsize{\fsize}{#1\fsize}\selectfont}%
  \ifx\svgwidth\undefined%
    \setlength{\unitlength}{377.94456482bp}%
    \ifx\svgscale\undefined%
      \relax%
    \else%
      \setlength{\unitlength}{\unitlength * \real{\svgscale}}%
    \fi%
  \else%
    \setlength{\unitlength}{\svgwidth}%
  \fi%
  \global\let\svgwidth\undefined%
  \global\let\svgscale\undefined%
  \makeatother%
  \begin{picture}(1,0.34530263)%
    \lineheight{1}%
    \setlength\tabcolsep{0pt}%
    \put(0.104712,0.3292206){\color[rgb]{0,0,0}\makebox(0,0)[t]{\lineheight{1.25}\smash{\begin{tabular}[t]{c}Input\end{tabular}}}}%
    \put(0.26838603,0.3292206){\color[rgb]{0,0,0}\makebox(0,0)[t]{\lineheight{1.25}\smash{\begin{tabular}[t]{c}ONet\end{tabular}}}}%
    \put(0.43206006,0.3292206){\color[rgb]{0,0,0}\makebox(0,0)[t]{\lineheight{1.25}\smash{\begin{tabular}[t]{c}ConvONet\end{tabular}}}}%
    \put(0.59573409,0.3292206){\color[rgb]{0,0,0}\makebox(0,0)[t]{\lineheight{1.25}\smash{\begin{tabular}[t]{c}IFNet\end{tabular}}}}%
    \put(0.75940817,0.3292206){\color[rgb]{0,0,0}\makebox(0,0)[t]{\lineheight{1.25}\smash{\begin{tabular}[t]{c}Ours\end{tabular}}}}%
    \put(0.92308221,0.3292206){\color[rgb]{0,0,0}\makebox(0,0)[t]{\lineheight{1.25}\smash{\begin{tabular}[t]{c}Ground Truth\end{tabular}}}}%
    \put(0.01608204,0.24059065){\color[rgb]{0,0,0}\rotatebox{90}{\makebox(0,0)[t]{\lineheight{0}\smash{\begin{tabular}[t]{c}Thingi10k\end{tabular}}}}}%
    \put(0.01608204,0.07691661){\color[rgb]{0,0,0}\rotatebox{90}{\makebox(0,0)[t]{\lineheight{0}\smash{\begin{tabular}[t]{c}ShapeNetV2\end{tabular}}}}}%
    \put(0,0){\includegraphics[width=\unitlength,page=1]{main_comparison_variable_vertex.pdf}}%
  \end{picture}%
\endgroup%

%% file: data/scripts/main_results.tex
\begin{tabular}{p{14mm}p{9mm}p{9mm}p{9mm}p{9mm}p{9mm}p{9mm}p{9mm}p{9mm}p{9mm}p{9mm}p{9mm}p{9mm}}
\toprule
 & \multicolumn{2}{c}{ONet} & \multicolumn{2}{c}{ConvONet} & \multicolumn{2}{c}{IFNet} & \multicolumn{2}{c}{POCO} & \multicolumn{2}{c}{3DS2VS} & \multicolumn{2}{c}{Ours} \\
\cmidrule(lr){2-3} \cmidrule(lr){4-5} \cmidrule(lr){6-7} \cmidrule(lr){8-9} \cmidrule(lr){10-11} \cmidrule(lr){12-13} 
 & CD$\downarrow$ \tiny  mean / std ($\cdot10^{-2}$) & NC$\uparrow$ \tiny  mean / std ($\cdot10^{-2}$) & CD$\downarrow$ \tiny  mean / std ($\cdot10^{-2}$) & NC$\uparrow$ \tiny  mean / std ($\cdot10^{-2}$) & CD$\downarrow$ \tiny  mean / std ($\cdot10^{-2}$) & NC$\uparrow$ \tiny  mean / std ($\cdot10^{-2}$) & CD$\downarrow$ \tiny  mean / std ($\cdot10^{-2}$) & NC$\uparrow$ \tiny  mean / std ($\cdot10^{-2}$) & CD$\downarrow$ \tiny  mean / std ($\cdot10^{-2}$) & NC$\uparrow$ \tiny  mean / std ($\cdot10^{-2}$) & CD$\downarrow$ \tiny  mean / std ($\cdot10^{-2}$) & NC$\uparrow$ \tiny  mean / std ($\cdot10^{-2}$) \\
\midrule
Dragon & 5.581\,/ 2.408 & 82.20\,/ 8.904 & 3.960\,/ 1.788 & 88.89\,/ 7.987 & 2.681\,/ 0.305 & 95.63\,/ 1.480 & 2.981\,/ 0.415 & 92.39\,/ 3.223 & 2.823\,/ 0.502 & 92.64\,/ 4.581 & \textbf{2.455}\,/ \textbf{0.168} & \textbf{96.25}\,/ \textbf{0.754} \\
\midrule
Armadillo & 4.451\,/ 2.667 & 83.35\,/ 6.613 & 2.377\,/ 1.840 & 90.80\,/ 5.827 & 1.259\,/ 1.678 & 95.87\,/ 2.529 & 2.594\,/ 0.387 & 89.06\,/ 2.636 & 1.291\,/ 0.286 & 94.12\,/ 2.858 & \textbf{0.956}\,/ \textbf{0.063} & \textbf{97.86}\,/ \textbf{0.591} \\
\midrule
DFaust & 16.50\,/ 3.618 & 82.70\,/ 4.671 & 1.627\,/ 1.769 & 94.50\,/ 3.210 & 1.015\,/ 2.671 & 96.52\,/ 3.158 & 3.480\,/ 0.269\footnotemark[1] & 84.93\,/ 1.261\footnotemark[1] & 1.181\,/ 1.519 & 95.33\,/ 2.170 & \textbf{0.627}\,/ \textbf{0.115} & \textbf{97.65}\,/ \textbf{0.602} \\
\midrule
Thingi10k & 12.69\,/ 5.523 & 62.16\,/ 11.01 & 10.62\,/ 4.349 & 67.11\,/ 13.09 & 3.888\,/ \textbf{1.057} & 92.27\,/ 5.171 & n\,/ a & n\,/ a & n\,/ a & n\,/ a & \textbf{3.420}\,/ 1.069 & \textbf{92.86}\,/ \textbf{4.830} \\
\midrule
ShapeNet v2 (w/ flip) & 34.89\,/ 3.142 & 59.61\,/ 4.807 & 3.734\,/ 1.933 & 75.43\,/ 6.454 & 2.180\,/ 0.513 & 82.34\,/ 3.942 & n\,/ a & n\,/ a & n\,/ a & n\,/ a & \textbf{2.011}\,/ \textbf{0.409} & \textbf{86.11}\,/ \textbf{3.731} \\
\midrule
% ShapeNet v2 (w/o flip) & 10.02\,/ 4.116 & 58.11\,/ 4.148 & 4.424\,/ 1.969 & 72.31\,/ 5.290 & \textbf{2.724}\,/ \textbf{0.523} & \textbf{81.61}\,/ \textbf{3.709} & n\,/ a & n\,/ a & n\,/ a & n\,/ a & 5.237\,/ 1.982 & 73.06\,/ 5.870 \\
\textcolor{lightgray}{ShapeNet v2 (w/o flip)} & \textcolor{lightgray}{10.02\,/ 4.116} & \textcolor{lightgray}{58.11\,/ 4.148} & \textcolor{lightgray}{4.424\,/ 1.969} & \textcolor{lightgray}{72.31\,/ 5.290} & \textcolor{lightgray}{\textbf{2.724}\,/ \textbf{0.523}} & \textcolor{lightgray}{\textbf{81.61}\,/ \textbf{3.709}} & \textcolor{lightgray}{n\,/ a} & \textcolor{lightgray}{n\,/ a} & \textcolor{lightgray}{n\,/ a} & \textcolor{lightgray}{n\,/ a} & \textcolor{lightgray}{5.237\,/ 1.982} & \textcolor{lightgray}{73.06\,/ 5.870} \\
\midrule

%% file: data/scripts/normals_comparison.tex
\begin{tabular}{lp{4.7mm}p{4.7mm}p{4.7mm}p{4.7mm}p{4.7mm}p{4.7mm}p{4.7mm}p{4.7mm}}
\toprule
 & \multicolumn{2}{c}{Pool w/o n} & \multicolumn{2}{c}{Proj. w/o n} & \multicolumn{2}{c}{Pool w/ n} & \multicolumn{2}{c}{Proj. w/ n} \\
\cmidrule(lr){2-3} \cmidrule(lr){4-5} \cmidrule(lr){6-7} \cmidrule(lr){8-9} 
 & CD$\downarrow$ \tiny  mean / std & NC$\uparrow$ \tiny  mean / std & CD$\downarrow$ \tiny  mean / std & NC$\uparrow$ \tiny  mean / std & CD$\downarrow$ \tiny  mean / std & NC$\uparrow$ \tiny  mean / std & CD$\downarrow$ \tiny  mean / std & NC$\uparrow$ \tiny  mean / std \\
\midrule
Dragon & 2.455\,/ 0.168 & 96.25\,/ 0.754 & \textbf{2.406}\,/ 0.161 & 96.33\,/ 0.473 & 2.486\,/ 0.186 & 96.37\,/ 0.374 & 2.427\,/ \textbf{0.148} & \textbf{96.53}\,/ \textbf{0.294} \\
\midrule
Armadillo & 0.956\,/ 0.063 & 97.86\,/ 0.591 & 0.934\,/ 0.048 & 98.22\,/ 0.519 & 0.934\,/ 0.047 & 98.27\,/ 0.451 & \textbf{0.906}\,/ \textbf{0.043} & \textbf{98.47}\,/ \textbf{0.354} \\
\midrule
DFaust & 0.627\,/ 0.115 & 97.65\,/ 0.602 & 0.662\,/ \textbf{0.082} & 97.41\,/ 0.534 & 0.700\,/ 0.149 & 97.61\,/ 0.592 & \textbf{0.626}\,/ 0.094 & \textbf{97.83}\,/ \textbf{0.492} \\
\midrule
Thingi10k & 3.420\,/ 1.069 & 92.86\,/ 4.830 & 3.222\,/ 0.829 & 94.26\,/ 3.904 & 3.446\,/ 0.917 & 95.53\,/ 2.617 & \textbf{3.075}\,/ \textbf{0.804} & \textbf{95.89}\,/ \textbf{2.384} \\
\midrule
ShapeNet & 2.011\,/ 0.409 & 86.11\,/ 3.731 & \textbf{1.888}\,/ \textbf{0.404} & \textbf{86.75}\,/ \textbf{3.595} & n\,/ a & n\,/ a & n\,/ a & n\,/ a \\
\midrule
Objaverse & 1.222\,/ \textbf{0.174} & 92.95\,/ 6.108 & 1.363\,/ 0.296 & 92.00\,/ 6.829 & \textbf{1.118}\,/ 0.237 & 94.87\,/ 4.474 & 1.151\,/ 0.565 & \textbf{95.02}\,/ \textbf{4.399} \\
\midrule

%% file: figures/comparison/pool_vs_proj_single.pdf_tex
%% Creator: Inkscape 1.3.2 (091e20e, 2023-11-25), www.inkscape.org
%% Accompanies image file 'pool_vs_proj_single.pdf' (pdf, eps, ps)
\begingroup%
  \makeatletter%
  \providecommand\color[2][]{%
    \errmessage{(Inkscape) Color is used for the text in Inkscape, but the package 'color.sty' is not loaded}%
    \renewcommand\color[2][]{}%
  }%
  \providecommand\transparent[1]{%
    \errmessage{(Inkscape) Transparency is used (non-zero) for the text in Inkscape, but the package 'transparent.sty' is not loaded}%
    \renewcommand\transparent[1]{}%
  }%
  \providecommand\rotatebox[2]{#2}%
  \newcommand*\fsize{\dimexpr\f@size pt\relax}%
  \newcommand*\lineheight[1]{\fontsize{\fsize}{#1\fsize}\selectfont}%
  \ifx\svgwidth\undefined%
    \setlength{\unitlength}{243.14749146bp}%
    \ifx\svgscale\undefined%
      \relax%
    \else%
      \setlength{\unitlength}{\unitlength * \real{\svgscale}}%
    \fi%
  \else%
    \setlength{\unitlength}{\svgwidth}%
  \fi%
  \global\let\svgwidth\undefined%
  \global\let\svgscale\undefined%
  \makeatother%
  \begin{picture}(1,1.08987347)%
    \lineheight{1}%
    \setlength\tabcolsep{0pt}%
    \put(0,0){\includegraphics[width=\unitlength,page=1]{pool_vs_proj_single.pdf}}%
    \put(0.05429214,0.97975512){\rotatebox{90}{\makebox(0,0)[t]{\lineheight{1.25}\smash{\begin{tabular}[t]{c}GT\end{tabular}}}}}%
    \put(0.04938359,0.75946494){\rotatebox{90}{\makebox(0,0)[t]{\lineheight{1.25}\smash{\begin{tabular}[t]{c}Proj.  w/ n\end{tabular}}}}}%
    \put(0.04938359,0.31714293){\rotatebox{90}{\makebox(0,0)[t]{\lineheight{1.25}\smash{\begin{tabular}[t]{c}Proj.\end{tabular}}}}}%
    \put(0.05510647,0.09637772){\rotatebox{90}{\makebox(0,0)[t]{\lineheight{1.25}\smash{\begin{tabular}[t]{c}Pool\end{tabular}}}}}%
    \put(0.05510647,0.53840571){\rotatebox{90}{\makebox(0,0)[t]{\lineheight{1.25}\smash{\begin{tabular}[t]{c}Pool w/ n\end{tabular}}}}}%
  \end{picture}%
\endgroup%

%% file: sections/5-conclusion.tex
\section{Conclusion and Future Work}
\label{sec:conclusion}
We have shown that our hybrid approach of interleaving graph and grid convolutions, including back and forth transfers, is capable of outperforming other state-of-the-art encoders on the surface reconstruction task from point cloud inputs.
We are able to do this using only surface information and additional unlabeled samples in the surrounding volume, which prevents us from having to solve for a ground truth signed distance field in advance.
We have also introduced a physically-inspired projection operator for particle-to-grid feature transfers in neural networks and shown its effectiveness.
Using this projection enables the reconstruction of both more detailed and less noisy surfaces.

We believe that this work could be foundational for further research regarding efficient models for 3D shape encoding.
For instance, exploring the sparsity of our latent grids could enable the usage of ever higher latent resolutions, resulting in accurate encoding of almost arbitrarily complex shapes.
Combined with small latent sizes, this could result in a more scalable architecture, where the accuracy can be determined by available compute capabilities.
Another interesting direction of further study is the use of the latent grid as a basis for editable neural fields.
Multiple latent representations could be blended, or operators could be learned on the latent codes to achieve specific effects.
Finally, in the spirit of using traditional geometric techniques within neural architectures, it could be explored whether projecting surface features to outer grid cells using methods such as fast marching or Laplacian smoothing could further improve predictions at distant points.

Code and data will be made available on our project website after publication.

%% file: paper.bbl
\begin{thebibliography}{47}
\providecommand{\natexlab}[1]{#1}
\providecommand{\url}[1]{\texttt{#1}}
\expandafter\ifx\csname urlstyle\endcsname\relax
  \providecommand{\doi}[1]{doi: #1}\else
  \providecommand{\doi}{doi: \begingroup \urlstyle{rm}\Url}\fi

\bibitem[Atzmon and Lipman(2020{\natexlab{a}})]{atzmonSALDSignAgnostic2020}
Matan Atzmon and Yaron Lipman.
\newblock {{SALD}}: {{Sign Agnostic Learning}} with {{Derivatives}}.
\newblock In \emph{9th {{International Conference}} on {{Learning
  Representations}}, {{ICLR}} 2021}, October 2020{\natexlab{a}}.

\bibitem[Atzmon and Lipman(2020{\natexlab{b}})]{atzmonSALSignAgnostic2020}
Matan Atzmon and Yaron Lipman.
\newblock {{SAL}}: {{Sign Agnostic Learning}} of {{Shapes From Raw Data}}.
\newblock In \emph{Proceedings of the {{IEEE}}/{{CVF Conference}} on {{Computer
  Vision}} and {{Pattern Recognition}}}, pages 2565--2574, 2020{\natexlab{b}}.

\bibitem[Bogo et~al.(2017)Bogo, Romero, {Pons-Moll}, and
  Black]{bogoDynamicFAUSTRegistering2017}
Federica Bogo, Javier Romero, Gerard {Pons-Moll}, and Michael~J. Black.
\newblock Dynamic {{FAUST}}: {{Registering Human Bodies}} in {{Motion}}.
\newblock In \emph{2017 {{IEEE Conference}} on {{Computer Vision}} and
  {{Pattern Recognition}} ({{CVPR}})}, pages 5573--5582. IEEE Computer Society,
  July 2017.
\newblock ISBN 978-1-5386-0457-1.
\newblock \doi{10.1109/CVPR.2017.591}.

\bibitem[Boulch and Marlet(2022)]{boulchPOCOPointConvolution2022}
Alexandre Boulch and Renaud Marlet.
\newblock {{POCO}}: {{Point Convolution}} for {{Surface Reconstruction}}.
\newblock In \emph{Proceedings of the {{IEEE}}/{{CVF Conference}} on {{Computer
  Vision}} and {{Pattern Recognition}}}, pages 6302--6314, 2022.

\bibitem[Buterez et~al.(2022)Buterez, Janet, Kiddle, Oglic, and
  Li{\`o}]{buterezGraphNeuralNetworks2022}
David Buterez, Jon~Paul Janet, Steven~J. Kiddle, Dino Oglic, and Pietro
  Li{\`o}.
\newblock Graph {{Neural Networks}} with {{Adaptive Readouts}}.
\newblock \emph{Advances in Neural Information Processing Systems},
  35:\penalty0 19746--19758, December 2022.

\bibitem[Chabra et~al.(2020)Chabra, Lenssen, Ilg, Schmidt, Straub, Lovegrove,
  and Newcombe]{chabraDeepLocalShapes2020}
Rohan Chabra, Jan~E. Lenssen, Eddy Ilg, Tanner Schmidt, Julian Straub, Steven
  Lovegrove, and Richard Newcombe.
\newblock Deep {{Local Shapes}}: {{Learning Local SDF Priors}} for {{Detailed
  3D Reconstruction}}.
\newblock In Andrea Vedaldi, Horst Bischof, Thomas Brox, and Jan-Michael Frahm,
  editors, \emph{Computer {{Vision}} -- {{ECCV}} 2020}, Lecture {{Notes}} in
  {{Computer Science}}, pages 608--625, Cham, 2020. Springer International
  Publishing.
\newblock ISBN 978-3-030-58526-6.
\newblock \doi{10.1007/978-3-030-58526-6_36}.

\bibitem[Chang et~al.(2015)Chang, Funkhouser, Guibas, Hanrahan, Huang, Li,
  Savarese, Savva, Song, Su, Xiao, Yi, and
  Yu]{changShapeNetInformationrich3D2015}
Angel~X. Chang, Thomas Funkhouser, Leonidas Guibas, Pat Hanrahan, Qixing Huang,
  Zimo Li, Silvio Savarese, Manolis Savva, Shuran Song, Hao Su, Jianxiong Xiao,
  Li~Yi, and Fisher Yu.
\newblock {{ShapeNet}}: {{An}} information-rich {{3D}} model repository.
\newblock Technical Report arXiv:1512.03012 [cs.GR], Stanford University ---
  Princeton University --- Toyota Technological Institute at Chicago, 2015.

\bibitem[Chen et~al.(2023)Chen, Xiang, Cho, Chang, Pershing, Maia, Chiaramonte,
  Carlberg, and Grinspun]{chenCROMContinuousReducedorder2023}
Peter~Yichen Chen, Jinxu Xiang, Dong~Heon Cho, Yue Chang, G~A Pershing,
  Henrique~Teles Maia, Maurizio~M Chiaramonte, Kevin~Thomas Carlberg, and Eitan
  Grinspun.
\newblock {{CROM}}: {{Continuous}} reduced-order modeling of {{PDEs}} using
  implicit neural representations.
\newblock In \emph{The Eleventh International Conference on Learning
  Representations}, 2023.

\bibitem[Chibane et~al.(2020{\natexlab{a}})Chibane, Alldieck, and
  {Pons-Moll}]{chibaneImplicitFunctionsFeature2020}
Julian Chibane, Thiemo Alldieck, and Gerard {Pons-Moll}.
\newblock Implicit {{Functions}} in {{Feature Space}} for {{3D Shape
  Reconstruction}} and {{Completion}}.
\newblock In \emph{Proceedings of the {{IEEE}}/{{CVF Conference}} on {{Computer
  Vision}} and {{Pattern Recognition}}}, pages 6970--6981, 2020{\natexlab{a}}.

\bibitem[Chibane et~al.(2020{\natexlab{b}})Chibane, {mir}, and
  {Pons-Moll}]{chibaneNeuralUnsignedDistance2020}
Julian Chibane, Mohamad~Aymen {mir}, and Gerard {Pons-Moll}.
\newblock Neural {{Unsigned Distance Fields}} for {{Implicit Function
  Learning}}.
\newblock In \emph{Advances in {{Neural Information Processing Systems}}},
  volume~33, pages 21638--21652. Curran Associates, Inc., 2020{\natexlab{b}}.

\bibitem[Deitke et~al.(2023)Deitke, Schwenk, Salvador, Weihs, Michel,
  VanderBilt, Schmidt, Ehsani, Kembhavi, and
  Farhadi]{deitkeObjaverseUniverseAnnotated2023}
Matt Deitke, Dustin Schwenk, Jordi Salvador, Luca Weihs, Oscar Michel, Eli
  VanderBilt, Ludwig Schmidt, Kiana Ehsani, Aniruddha Kembhavi, and Ali
  Farhadi.
\newblock Objaverse: {{A Universe}} of {{Annotated 3D Objects}}.
\newblock In \emph{Proceedings of the {{IEEE}}/{{CVF Conference}} on {{Computer
  Vision}} and {{Pattern Recognition}}}, pages 13142--13153, 2023.

\bibitem[Fey and Lenssen(2019)]{feyFastGraphRepresentation2019}
Matthias Fey and Jan~E. Lenssen.
\newblock Fast graph representation learning with {{PyTorch Geometric}}.
\newblock In \emph{{{ICLR}} Workshop on Representation Learning on Graphs and
  Manifolds}, 2019.

\bibitem[Gropp et~al.(2020)Gropp, Yariv, Haim, Atzmon, and
  Lipman]{groppImplicitGeometricRegularization2020}
Amos Gropp, Lior Yariv, Niv Haim, Matan Atzmon, and Yaron Lipman.
\newblock Implicit {{Geometric Regularization}} for {{Learning Shapes}}.
\newblock In \emph{Proceedings of the 37th {{International Conference}} on
  {{Machine Learning}}}, pages 3789--3799. PMLR, November 2020.

\bibitem[Harlow(1962)]{harlowParticleincellMethodNumerical1962}
Francis Harlow.
\newblock The particle-in-cell method for numerical solution of problems in
  fluid dynamics.
\newblock Technical Report LADC-5288, 4769185, March 1962.

\bibitem[Hertz et~al.(2021)Hertz, Perel, Giryes, {Sorkine-Hornung}, and
  {Cohen-Or}]{hertzSAPESpatiallyAdaptiveProgressive2021}
Amir Hertz, Or~Perel, Raja Giryes, Olga {Sorkine-Hornung}, and Daniel
  {Cohen-Or}.
\newblock {{SAPE}}: {{Spatially-Adaptive Progressive Encoding}} for {{Neural
  Optimization}}.
\newblock In \emph{Advances in {{Neural Information Processing Systems}}},
  volume~34, pages 8820--8832. Curran Associates, Inc., 2021.

\bibitem[Hu et~al.(2018)Hu, Zhou, Gao, Jacobson, Zorin, and
  Panozzo]{huTetrahedralMeshingWild2018}
Yixin Hu, Qingnan Zhou, Xifeng Gao, Alec Jacobson, Denis Zorin, and Daniele
  Panozzo.
\newblock Tetrahedral meshing in the wild.
\newblock \emph{ACM Transactions on Graphics}, 37\penalty0 (4):\penalty0
  60:1--60:14, July 2018.
\newblock ISSN 0730-0301.
\newblock \doi{10.1145/3197517.3201353}.

\bibitem[Jiang et~al.(2015)Jiang, Schroeder, Selle, Teran, and
  Stomakhin]{jiangAffineParticleincellMethod2015}
Chenfanfu Jiang, Craig Schroeder, Andrew Selle, Joseph Teran, and Alexey
  Stomakhin.
\newblock The affine particle-in-cell method.
\newblock \emph{ACM Transactions on Graphics}, 34\penalty0 (4):\penalty0
  51:1--51:10, July 2015.
\newblock ISSN 0730-0301.
\newblock \doi{10.1145/2766996}.

\bibitem[Jiang et~al.(2020)Jiang, Sud, Makadia, Huang, Nie{\ss}ner, and
  Funkhouser]{jiangLocalImplicitGrid2020}
Chiyu Jiang, Avneesh Sud, Ameesh Makadia, Jingwei Huang, Matthias Nie{\ss}ner,
  and Thomas Funkhouser.
\newblock Local {{Implicit Grid Representations}} for {{3D Scenes}}.
\newblock In \emph{2020 {{IEEE}}/{{CVF Conference}} on {{Computer Vision}} and
  {{Pattern Recognition}} ({{CVPR}})}, pages 6000--6009. IEEE Computer Society,
  June 2020.
\newblock ISBN 978-1-72817-168-5.
\newblock \doi{10.1109/CVPR42600.2020.00604}.

\bibitem[Li et~al.(2020)Li, Ferguson, Schneider, Langlois, Zorin, Panozzo,
  Jiang, and Kaufman]{liIncrementalPotentialContact2020}
Minchen Li, Zachary Ferguson, Teseo Schneider, Timothy Langlois, Denis Zorin,
  Daniele Panozzo, Chenfanfu Jiang, and Danny~M. Kaufman.
\newblock Incremental potential contact: Intersection-and inversion-free,
  large-deformation dynamics.
\newblock \emph{ACM Transactions on Graphics}, 39\penalty0 (4):\penalty0
  49:49:1--49:49:20, August 2020.
\newblock ISSN 0730-0301.
\newblock \doi{10.1145/3386569.3392425}.

\bibitem[Liu et~al.(2019)Liu, Tang, Lin, and
  Han]{liuPointVoxelCNNEfficient2019}
Zhijian Liu, Haotian Tang, Yujun Lin, and Song Han.
\newblock Point-{{Voxel CNN}} for {{Efficient 3D Deep Learning}}.
\newblock In \emph{Advances in {{Neural Information Processing Systems}}},
  volume~32. Curran Associates, Inc., 2019.

\bibitem[Lombardi et~al.(2019)Lombardi, Simon, Saragih, Schwartz, Lehrmann, and
  Sheikh]{lombardiNeuralVolumesLearning2019}
Stephen Lombardi, Tomas Simon, Jason Saragih, Gabriel Schwartz, Andreas
  Lehrmann, and Yaser Sheikh.
\newblock Neural volumes: Learning dynamic renderable volumes from images.
\newblock \emph{ACM Transactions on Graphics}, 38\penalty0 (4):\penalty0
  65:1--65:14, July 2019.
\newblock ISSN 0730-0301.
\newblock \doi{10.1145/3306346.3323020}.

\bibitem[Long et~al.(2023)Long, Lin, Liu, Liu, Wang, Theobalt, Komura, and
  Wang]{longNeuralUDFLearningUnsigned2023}
Xiaoxiao Long, Cheng Lin, Lingjie Liu, Yuan Liu, Peng Wang, Christian Theobalt,
  Taku Komura, and Wenping Wang.
\newblock {{NeuralUDF}}: {{Learning Unsigned Distance Fields}} for {{Multi-View
  Reconstruction}} of {{Surfaces With Arbitrary Topologies}}.
\newblock In \emph{Proceedings of the {{IEEE}}/{{CVF Conference}} on {{Computer
  Vision}} and {{Pattern Recognition}}}, pages 20834--20843, 2023.

\bibitem[Ma et~al.(2021)Ma, Han, Liu, and
  Zwicker]{maNeuralPullLearningSigned2021}
Baorui Ma, Zhizhong Han, Yu-Shen Liu, and Matthias Zwicker.
\newblock Neural-{{Pull}}: {{Learning Signed Distance Function}} from {{Point}}
  clouds by {{Learning}} to {{Pull Space}} onto {{Surface}}.
\newblock In \emph{Proceedings of the 38th {{International Conference}} on
  {{Machine Learning}}}, pages 7246--7257. PMLR, July 2021.

\bibitem[Mehta et~al.(2021)Mehta, Gharbi, Barnes, Shechtman, Ramamoorthi, and
  Chandraker]{mehtaModulatedPeriodicActivations2021}
Ishit Mehta, Micha{\"e}l Gharbi, Connelly Barnes, Eli Shechtman, Ravi
  Ramamoorthi, and Manmohan Chandraker.
\newblock Modulated {{Periodic Activations}} for {{Generalizable Local
  Functional Representations}}.
\newblock In \emph{2021 {{IEEE}}/{{CVF International Conference}} on {{Computer
  Vision}} ({{ICCV}})}, pages 14194--14203, Los Alamitos, CA, USA, April 2021.
  arXiv.
\newblock \doi{10.1109/ICCV48922.2021.01395}.

\bibitem[Mescheder et~al.(2019)Mescheder, Oechsle, Niemeyer, Nowozin, and
  Geiger]{meschederOccupancyNetworksLearning2019}
Lars Mescheder, Michael Oechsle, Michael Niemeyer, Sebastian Nowozin, and
  Andreas Geiger.
\newblock Occupancy {{Networks}}: {{Learning 3D Reconstruction}} in {{Function
  Space}}.
\newblock In \emph{2019 {{IEEE}}/{{CVF Conference}} on {{Computer Vision}} and
  {{Pattern Recognition}} ({{CVPR}})}, pages 4455--4465, June 2019.
\newblock \doi{10.1109/CVPR.2019.00459}.

\bibitem[Mildenhall et~al.(2020)Mildenhall, Srinivasan, Tancik, Barron,
  Ramamoorthi, and Ng]{mildenhallNeRFRepresentingScenes2020}
Ben Mildenhall, Pratul~P. Srinivasan, Matthew Tancik, Jonathan~T. Barron, Ravi
  Ramamoorthi, and Ren Ng.
\newblock {{NeRF}}: {{Representing Scenes}} as {{Neural Radiance Fields}} for
  {{View Synthesis}}.
\newblock In Andrea Vedaldi, Horst Bischof, Thomas Brox, and Jan-Michael Frahm,
  editors, \emph{Computer {{Vision}} -- {{ECCV}} 2020}, Lecture {{Notes}} in
  {{Computer Science}}, pages 405--421, Cham, 2020. Springer International
  Publishing.
\newblock ISBN 978-3-030-58452-8.
\newblock \doi{10.1007/978-3-030-58452-8_24}.

\bibitem[Ouasfi and Boukhayma(2022)]{ouasfiFewZeroLevel2022}
Amine Ouasfi and Adnane Boukhayma.
\newblock Few `{{Zero Level Set}}'-{{Shot Learning}} of~{{Shape Signed Distance
  Functions}} in~{{Feature Space}}.
\newblock In Shai Avidan, Gabriel Brostow, Moustapha Ciss{\'e}, Giovanni~Maria
  Farinella, and Tal Hassner, editors, \emph{Computer {{Vision}} -- {{ECCV}}
  2022}, Lecture {{Notes}} in {{Computer Science}}, pages 561--578, Cham, 2022.
  Springer Nature Switzerland.
\newblock ISBN 978-3-031-19824-3.
\newblock \doi{10.1007/978-3-031-19824-3_33}.

\bibitem[Park et~al.(2019)Park, Florence, Straub, Newcombe, and
  Lovegrove]{parkDeepSDFLearningContinuous2019}
Jeong~Joon Park, Peter Florence, Julian Straub, Richard Newcombe, and Steven
  Lovegrove.
\newblock {{DeepSDF}}: {{Learning Continuous Signed Distance Functions}} for
  {{Shape Representation}}.
\newblock In \emph{2019 {{IEEE}}/{{CVF Conference}} on {{Computer Vision}} and
  {{Pattern Recognition}} ({{CVPR}})}, pages 165--174, June 2019.
\newblock \doi{10.1109/CVPR.2019.00025}.

\bibitem[Paszke et~al.(2019)Paszke, Gross, Massa, Lerer, Bradbury, Chanan,
  Killeen, Lin, Gimelshein, Antiga, Desmaison, Kopf, Yang, DeVito, Raison,
  Tejani, Chilamkurthy, Steiner, Fang, Bai, and
  Chintala]{paszkePyTorchImperativeStyle2019}
Adam Paszke, Sam Gross, Francisco Massa, Adam Lerer, James Bradbury, Gregory
  Chanan, Trevor Killeen, Zeming Lin, Natalia Gimelshein, Luca Antiga, Alban
  Desmaison, Andreas Kopf, Edward Yang, Zachary DeVito, Martin Raison, Alykhan
  Tejani, Sasank Chilamkurthy, Benoit Steiner, Lu~Fang, Junjie Bai, and Soumith
  Chintala.
\newblock {{PyTorch}}: {{An Imperative Style}}, {{High-Performance Deep
  Learning Library}}.
\newblock In \emph{Advances in {{Neural Information Processing Systems}}},
  volume~32. Curran Associates, Inc., 2019.

\bibitem[Peng et~al.(2020)Peng, Niemeyer, Mescheder, Pollefeys, and
  Geiger]{pengConvolutionalOccupancyNetworks2020}
Songyou Peng, Michael Niemeyer, Lars Mescheder, Marc Pollefeys, and Andreas
  Geiger.
\newblock Convolutional {{Occupancy Networks}}.
\newblock In Andrea Vedaldi, Horst Bischof, Thomas Brox, and Jan-Michael Frahm,
  editors, \emph{Computer {{Vision}} -- {{ECCV}} 2020}, Lecture {{Notes}} in
  {{Computer Science}}, pages 523--540, Cham, 2020. Springer International
  Publishing.
\newblock ISBN 978-3-030-58580-8.
\newblock \doi{10.1007/978-3-030-58580-8_31}.

\bibitem[Peng et~al.(2021)Peng, Jiang, Liao, Niemeyer, Pollefeys, and
  Geiger]{pengShapePointsDifferentiable2021}
Songyou Peng, Chiyu Jiang, Yiyi Liao, Michael Niemeyer, Marc Pollefeys, and
  Andreas Geiger.
\newblock Shape {{As Points}}: {{A Differentiable Poisson Solver}}.
\newblock In \emph{Advances in {{Neural Information Processing Systems}}},
  volume~34, pages 13032--13044. Curran Associates, Inc., 2021.

\bibitem[Qi et~al.(2017)Qi, Su, Mo, and Guibas]{qiPointNetDeepLearning2017}
Charles~R. Qi, Hao Su, Kaichun Mo, and Leonidas~J. Guibas.
\newblock {{PointNet}}: {{Deep Learning}} on {{Point Sets}} for {{3D
  Classification}} and {{Segmentation}}.
\newblock In \emph{Proceedings of the {{IEEE Conference}} on {{Computer
  Vision}} and {{Pattern Recognition}}}, pages 652--660, 2017.

\bibitem[Rahaman et~al.(2019)Rahaman, Baratin, Arpit, Draxler, Lin, Hamprecht,
  Bengio, and Courville]{rahamanSpectralBiasNeural2019}
Nasim Rahaman, Aristide Baratin, Devansh Arpit, Felix Draxler, Min Lin, Fred
  Hamprecht, Yoshua Bengio, and Aaron Courville.
\newblock On the {{Spectral Bias}} of {{Neural Networks}}.
\newblock In \emph{Proceedings of the 36th {{International Conference}} on
  {{Machine Learning}}}, pages 5301--5310. PMLR, May 2019.

\bibitem[Raissi et~al.(2019)Raissi, Perdikaris, and
  Karniadakis]{raissiPhysicsinformedNeuralNetworks2019}
M.~Raissi, P.~Perdikaris, and G.~E. Karniadakis.
\newblock Physics-informed neural networks: {{A}} deep learning framework for
  solving forward and inverse problems involving nonlinear partial differential
  equations.
\newblock \emph{Journal of Computational Physics}, 378:\penalty0 686--707,
  February 2019.
\newblock ISSN 0021-9991.
\newblock \doi{10.1016/j.jcp.2018.10.045}.

\bibitem[Sitzmann et~al.(2020{\natexlab{a}})Sitzmann, Chan, Tucker, Snavely,
  and Wetzstein]{sitzmannMetaSDFMetalearningSigned2020}
Vincent Sitzmann, Eric Chan, Richard Tucker, Noah Snavely, and Gordon
  Wetzstein.
\newblock {{MetaSDF}}: {{Meta-Learning Signed Distance Functions}}.
\newblock In \emph{Advances in {{Neural Information Processing Systems}}},
  volume~33, pages 10136--10147. Curran Associates, Inc., 2020{\natexlab{a}}.

\bibitem[Sitzmann et~al.(2020{\natexlab{b}})Sitzmann, Martel, Bergman, Lindell,
  and Wetzstein]{sitzmannImplicitNeuralRepresentations2020}
Vincent Sitzmann, Julien Martel, Alexander Bergman, David Lindell, and Gordon
  Wetzstein.
\newblock Implicit {{Neural Representations}} with {{Periodic Activation
  Functions}}.
\newblock In \emph{Advances in {{Neural Information Processing Systems}}},
  volume~33, pages 7462--7473. Curran Associates, Inc., 2020{\natexlab{b}}.

\bibitem[Smith et~al.(2021)Smith, Azizzadenesheli, and
  Ross]{smithEikoNetSolvingEikonal2021}
Jonathan~D. Smith, Kamyar Azizzadenesheli, and Zachary~E. Ross.
\newblock {{EikoNet}}: {{Solving}} the {{Eikonal Equation With Deep Neural
  Networks}}.
\newblock \emph{IEEE Transactions on Geoscience and Remote Sensing},
  59\penalty0 (12):\penalty0 10685--10696, December 2021.
\newblock ISSN 1558-0644.
\newblock \doi{10.1109/TGRS.2020.3039165}.

\bibitem[Takikawa et~al.(2021)Takikawa, Litalien, Yin, Kreis, Loop,
  Nowrouzezahrai, Jacobson, McGuire, and
  Fidler]{takikawaNeuralGeometricLevel2021}
Towaki Takikawa, Joey Litalien, Kangxue Yin, Karsten Kreis, Charles Loop, Derek
  Nowrouzezahrai, Alec Jacobson, Morgan McGuire, and Sanja Fidler.
\newblock Neural {{Geometric Level}} of {{Detail}}: {{Real-time Rendering}}
  with {{Implicit 3D Shapes}}.
\newblock \emph{Proceedings of the IEEE/CVF Conference on Computer Vision and
  Pattern Recognition (CVPR)}, 2021.

\bibitem[Tancik et~al.(2020)Tancik, Srinivasan, Mildenhall, {Fridovich-Keil},
  Raghavan, Singhal, Ramamoorthi, Barron, and Ng]{tancikFourierFeaturesLet2020}
Matthew Tancik, Pratul~P. Srinivasan, Ben Mildenhall, Sara {Fridovich-Keil},
  Nithin Raghavan, Utkarsh Singhal, Ravi Ramamoorthi, Jonathan~T. Barron, and
  Ren Ng.
\newblock Fourier {{Features Let Networks Learn High Frequency Functions}} in
  {{Low Dimensional Domains}}.
\newblock In \emph{Proceedings of the 34th {{International Conference}} on
  {{Neural Information Processing Systems}}}, {{NIPS}}'20, Red Hook, NY, USA,
  2020. Curran Associates Inc.
\newblock ISBN 978-1-71382-954-6.

\bibitem[Tang et~al.(2021)Tang, Lei, Xu, Ma, Jia, and
  Zhang]{tangSAConvONetSignAgnosticOptimization2021}
Jiapeng Tang, Jiabao Lei, Dan Xu, Feiying Ma, Kui Jia, and Lei Zhang.
\newblock {{SA-ConvONet}}: {{Sign-Agnostic Optimization}} of {{Convolutional
  Occupancy Networks}}.
\newblock In \emph{Proceedings of the {{IEEE}}/{{CVF International Conference}}
  on {{Computer Vision}}}, pages 6504--6513, 2021.

\bibitem[Vaswani et~al.(2017)Vaswani, Shazeer, Parmar, Uszkoreit, Jones, Gomez,
  Kaiser, and Polosukhin]{vaswaniAttentionAllYou2017}
Ashish Vaswani, Noam Shazeer, Niki Parmar, Jakob Uszkoreit, Llion Jones,
  Aidan~N Gomez, {\L}ukasz Kaiser, and Illia Polosukhin.
\newblock Attention is all you need.
\newblock In I.~Guyon, U.~Von Luxburg, S.~Bengio, H.~Wallach, R.~Fergus,
  S.~Vishwanathan, and R.~Garnett, editors, \emph{Advances in Neural
  Information Processing Systems}, volume~30. Curran Associates, Inc., 2017.

\bibitem[Wang et~al.(2022)Wang, Rahmann, and
  {Sorkine-Hornung}]{wangGeometryconsistentNeuralShape2022}
Yifan Wang, Lukas Rahmann, and Olga {Sorkine-Hornung}.
\newblock Geometry-consistent neural shape representation with implicit
  displacement fields.
\newblock In \emph{The Tenth International Conference on Learning
  Representations}. OpenReview, 2022.

\bibitem[Wang et~al.(2019)Wang, Sun, Liu, Sarma, Bronstein, and
  Solomon]{wangDynamicGraphCNN2019}
Yue Wang, Yongbin Sun, Ziwei Liu, Sanjay~E. Sarma, Michael~M. Bronstein, and
  Justin~M. Solomon.
\newblock Dynamic {{Graph CNN}} for {{Learning}} on {{Point Clouds}}.
\newblock \emph{ACM Transactions on Graphics}, 38\penalty0 (5):\penalty0
  146:1--146:12, October 2019.
\newblock ISSN 0730-0301.
\newblock \doi{10.1145/3326362}.

\bibitem[Xie et~al.(2022)Xie, Takikawa, Saito, Litany, Yan, Khan, Tombari,
  Tompkin, Sitzmann, and Sridhar]{xieNeuralFieldsVisual2022}
Yiheng Xie, Towaki Takikawa, Shunsuke Saito, Or~Litany, Shiqin Yan, Numair
  Khan, Federico Tombari, James Tompkin, Vincent Sitzmann, and Srinath Sridhar.
\newblock Neural {{Fields}} in {{Visual Computing}} and {{Beyond}}.
\newblock \emph{Computer Graphics Forum}, 41\penalty0 (2):\penalty0 641--676,
  2022.
\newblock ISSN 1467-8659.
\newblock \doi{10.1111/cgf.14505}.

\bibitem[Xu et~al.(2019)Xu, Zhang, and Xiao]{xuTrainingBehaviorDeep2019}
Zhi-Qin~John Xu, Yaoyu Zhang, and Yanyang Xiao.
\newblock Training {{Behavior}} of {{Deep Neural Network}} in {{Frequency
  Domain}}.
\newblock In Tom Gedeon, Kok~Wai Wong, and Minho Lee, editors, \emph{Neural
  {{Information Processing}}}, Lecture {{Notes}} in {{Computer Science}}, pages
  264--274, Cham, 2019. Springer International Publishing.
\newblock ISBN 978-3-030-36708-4.
\newblock \doi{10.1007/978-3-030-36708-4_22}.

\bibitem[Zhang et~al.(2022)Zhang, Niessner, and
  Wonka]{zhang3DILGIrregularLatent2022}
Biao Zhang, Matthias Niessner, and Peter Wonka.
\newblock {{3DILG}}: {{Irregular Latent Grids}} for {{3D Generative Modeling}}.
\newblock \emph{Advances in Neural Information Processing Systems},
  35:\penalty0 21871--21885, December 2022.

\bibitem[Zhang et~al.(2023)Zhang, Tang, Nie{\ss}ner, and
  Wonka]{zhang3DShape2VecSet3DShape2023}
Biao Zhang, Jiapeng Tang, Matthias Nie{\ss}ner, and Peter Wonka.
\newblock {{3DShape2VecSet}}: {{A 3D Shape Representation}} for {{Neural
  Fields}} and {{Generative Diffusion Models}}.
\newblock \emph{ACM Transactions on Graphics}, 42\penalty0 (4):\penalty0
  92:1--92:16, July 2023.
\newblock ISSN 0730-0301.
\newblock \doi{10.1145/3592442}.

\end{thebibliography}
